# Title Page

Title:

Fuzzy Mutation Embedded Hybrids of Gravitational Search and Particle Swarm Optimization Methods for Engineering Design Problems

Authors:
Devroop Kar[a], Manosij Ghosh[a], Ritam Guha[a], Ram Sarkar[a], Laura García-Hernández[b], Ajith Abraham[c,d]

Affiliation:
[a]Computer Science and Engineering Department, Jadavpur University, 188, Raja S.C. Mallick Road, Kolkata – 700032, West Bengal, India. Kolkata, India
[b]Area of Project Engineering, Universidad de Córdoba, Spain
[c]Scientific Network for Innovation and Research Excellence, Machine Intelligence Research Labs (MIR Labs) Auburn, Washington 98071, USA
[d]Department of Computer Science, University of Pretoria, South Africa

E-mail: kardevroop@gmail.com, manosij1996@gmail.com, ritamguha16@gmail.com raamsarkar@gmail.com, ir1gahel@uco.es, ajith.abraham@ieee.org

Corresponding Author: Ritam Guha
Email: ritamguha16@gmail.com
Phone No.: +91-9831524527

Orchid ids:
Manosij Ghosh - 0000-0003-2954-9876
Ritam Guha - 0000-0002-1375-777X
Ram Sarkar - 0000-0001-8813-4086
Ajith Abraham - 0000-0002-0169-6738
Laura García-Hernández - 0000-0002-8394-5696

**Abstract:** Gravitational Search Algorithm (GSA) and Particle Swarm Optimization (PSO) are nature-inspired, swarm-based optimization algorithms respectively. Though they have been widely used for single-objective optimization since their inception, they suffer from premature convergence. Even though the hybrids of GSA and PSO perform much better, the problem remains. Hence, to solve this issue we have proposed a fuzzy mutation model for two hybrid versions of PSO and GSA – Gravitational Particle Swarm (GPS) and PSOGSA. The developed algorithms are called Mutation based GPS (MGPS) and Mutation based PSOGSA (MPSOGSA). The mutation operator is based on a fuzzy model where the probability of mutation has been calculated based on the closeness of particle to population centroid and improvement in the particle value. We have evaluated these two new algorithms on 23 benchmark functions of three categories (unimodal, multi-modal and multi-modal with fixed dimension). The experimental outcome shows that our proposed model outperforms their corresponding ancestors, MGPS outperforms GPS 13 out of 23 times (56.52%) and MPSOGSA outperforms PSOGSA 17 times out of 23 (73.91%). We have also compared our results against those of recent optimization algorithms such as Sine Cosine Algorithm (SCA), Opposition-Based SCA, and Volleyball Premier League Algorithm (VPL). In addition, we have applied our proposed algorithms on some classic engineering design problems and the outcomes are satisfactory. The related codes of the proposed algorithms can be found in this link: Fuzzy-Mutation-Embedded-Hybrids-of-GSA-and-PSO.

# Fuzzy Mutation embedded Hybrids of Gravitational Search and Particle Swarm Optimization Methods for Engineering Design Problems

**Abstract:** Gravitational Search Algorithm (GSA) and Particle Swarm Optimization (PSO) are nature-inspired, swarm-based optimization algorithms respectively. Though they have been widely used for single-objective optimization since their inception, they suffer from premature convergence. Even though the hybrids of GSA and PSO perform much better, the problem remains. Hence, to solve this issue we have proposed a fuzzy mutation model for two hybrid versions of PSO and GSA – Gravitational Particle Swarm (GPS) and PSOGSA. The developed algorithms are called Mutation based GPS (MGPS) and Mutation based PSOGSA (MPSOGSA). The mutation operator is based on a fuzzy model where the probability of mutation has been calculated based on the closeness of particle to population centroid and improvement in the particle value. We have evaluated these two new algorithms on 23 benchmark functions of three categories (unimodal, multi-modal and multi-modal with fixed dimension). The experimental outcome shows that our proposed model outperforms their corresponding ancestors, MGPS outperforms GPS 13 out of 23 times (56.52%) and MPSOGSA outperforms PSOGSA 17 times out of 23 (73.91%). We have also compared our results against those of recent optimization algorithms such as Sine Cosine Algorithm (SCA), Opposition-Based SCA, and Volleyball Premier League Algorithm (VPL). In addition, we have applied our proposed algorithms on some classic engineering design problems and the outcomes are satisfactory. The related codes of the proposed algorithms can be found in this link: Fuzzy-Mutation-Embedded-Hybrids-of-GSA-and-PSO.

## 1. Introduction

An optimization problem maximizes or minimizes a real function by systematically choosing input values from within an allowed set. This problem is of particular interest in the fields of operations research and certain engineering applications. There are major subfields to this section of mathematics including convex programming, stochastic programming and optimization, meta-heuristic programming, etc. While meta-heuristics do not guarantee that the best solution will be found, they are widely used to find good approximate (optimal) solutions for many complicated optimization problems. Gravitational Search Algorithm (GSA) [1], Particle Swarm Algorithm (PSO) [2], Genetic Algorithm [3], Cuckoo search algorithm [4], Grey Wolf Algorithm [5] are some of the effective meta-heuristic algorithms based on natural phenomena that have yielded promising results over the years. Of particular importance are GSA and PSO which are swarm-based meta-heuristics.

PSO was proposed by Ebelhart and Kennedy in 1995 [2]. It simulates the social behavior of birds and fish. Its ability to efficiently solve numerous scientific and engineering optimization problems has given it increasing support and acceptance among researchers. Apart from optimization problems, PSO has been applied in feature selection [6] [7] as well as data clustering [8]. The algorithm has been used in the field of electromagnetics [9] and image segmentation [10] [11] [12]. PSO has also been modified to improve its convergence capabilities to create a Quantum based PSO [13] as well as an adaptive version described in [14]. Oppositional learning has also been used to improve PSO's exploration capability by preventing the particles from getting trapped in a local minimum in OBPSO[15].

GSA was proposed by Rashedi and Saryazdi in 2008 [1] for solving single-objective optimization problems. It is based on Newtonian gravity stating "Every particle in the universe attracts every other particle with a force which is directly proportional to the product of their masses and inversely proportional to the square of the distance between the masses." Variations of GSA include QIGSA (Quantum Inspired GSA) [16] which uses quantum mechanics theories to prevent the premature convergence problem of GSA. CGSA [17] combines chaos with GSA for selection of parameters using chaos theory. A binary version of the GSA has also been developed called BGSA [18] by the same authors, Rashedi and Saryazdi that efficiently tackles feature selection and dimension reduction [19]. This algorithm has also been combined with *Simulated Annealing* to form GABSA (Gravitation Algorithm Based Simulated Annealing) [20]. GSA has been modified and used for various other real-world problems like data clustering [21]. GSA has also been implemented in the optimization of power despatch in a grid [16][22], forecasting turbine heat rate [23] and also in image segmentation [24].

Hybrid algorithms combining PSO and GSA have also been proposed in the literature. These include PSOGSA [25] which integrates the ability of exploitation in PSO with the exploration ability in GSA to harness both algorithms' strengths. Comparison of the hybrid algorithms with both the standard PSO and GSA algorithms by testing against some benchmark functions shows that the hybrid algorithm has a better capability to escape from local optima with faster convergence rate than the standard PSO and GSA. Another such hybrid algorithm is the Gravitational Particle Swarm (GPS) [26] in which a GPS agent has attributes of both GSA and PSO. GPS agents update their respective positions with PSO and GSA velocities. GPS agents, therefore, can exhibit both social and cognitive behavior, and motion of birds in flight as shown by the PSO algorithm [2] along with the law of gravity utilized in GSA[1]. Results show that both GPS and PSOGSA outperform both PSO and GSA by a significant margin. Therefore, due to these reasons, we choose these two hybrids of GSA and PSO as our base algorithm for further improvement.

But the main limitation of these hybrid algorithms is that they have poor local search capabilities especially in GPS which is pointed out in [27]. Due to their fast convergence rate, they suffer from premature convergence. So, the algorithm may in most cases converge to some local optima (sometimes very close to the global optima) and is stuck there which hinders the achievement of the best result, like in the case of their ancestors. To get rid of this problem, the concept of an exploratory operator called centroid based fuzzy mutation has been introduced in GPS and PSOGSA. This addition of mutation allows us to address the problem of premature convergence, which is the main contribution of this paper. We have tested the proposed algorithms on a set of benchmark functions and the results corroborate our assumption.

Apart from the algorithms mentioned previously, we have also compared our results to algorithms like Opposition-based PSO (OBPSO, 2007) [15], which was a step in the direction to address the tendency of PSO to get trapped in a local optima, as well as, Sine Cosine Algorithm(SCA, 2016) [28], Opposition-based SCA (OBSCA, 2017) [29], Social Spider Algorithm (SSO, 2013) [30], League Championship Algorithm (LCA, 2009) [31], Soccer League Competition Algorithm (SLC, 2014) [32] and Volleyball Premier League Algorithm (VPL, 2018) [33]. These algorithms are outperformed in almost 70% cases by the proposed algorithms.

Engineering design problems involve defining values of design parameters which gives the best output for a mechanical device, structure, or system. This process for determining the best values is called engineering optimization. Sometimes many variables need to be adjusted while satisfying several conflicting objectives and/or constraints. Therefore, implicitly determining values using intuition becomes very difficult. Optimization using evolutionary algorithms come into play here. There are several works on the use of evolutionary algorithms in engineering design problems [34], [35]. To portray the usefulness of our algorithms we have applied them to five benchmark engineering design problems [36] - Tension/Compression Spring Design problem, Gear train design problem, Welded Beam Design problem, Pressure design vessel problem and Closed coil helical spring design problem.

The contributions of this manuscript are presented below:

i. Development of an effective fuzzy-based mutation for hybrids of PSO and GSA namely GPS and PSOGSA.
ii. Evaluation of our algorithms on several benchmark functions to prove the effectiveness and relative superiority of the same.
iii. Application of our algorithms on some classical engineering design problems to show their practical usage.

## 2. Methods and Methodologies

The hybrids algorithms we consider in our work - PSOGSA and GPS are described briefly in sections 2.1 and 2.2 respectively. It should be noted that the points in our search space are referred to as points, particles and agents inter-changeably and refer to the same.

### 2.1 Hybrid Particle Swarm and Gravitational Search Algorithm (PSOGSA)

PSOGSA [25] is a novel hybrid optimization algorithm, combining the strengths of both PSO and GSA. It has been shown through results that this algorithm outperforms both PSO and GSA in terms of improved exploration and exploitation. The original version of this algorithm is well suited for problems with continuous search space.

The basic idea of PSOGSA is to combine the ability of social thinking ($gbest$) in PSO with exploration capability of GSA. The PSOGSA algorithm was mathematically modeled as similar to PSO and GSA, every search agent has a position vector reflecting the current position in search spaces as follows:

$$X_i = (x_{i1}, \dots, x_{id}, \dots, x_{iD}), i = 1,2, \dots, N \quad (1)$$

N is the number of search agents, $d$ is the index and $D$ is the dimension of the problem, and $x_{id}$ is the position of the $i^{th}$ agent in the $d^{th}$ dimension. Optimization process begins with filling out the position matrix with random values. During optimization, the gravitational force from agent $j$ on agent $i$ at a specific time $t$ is defined as follows:

$$F_{ij}^{d} = G(t) * \left(\frac{M_{pj}(t) * M_{aj}(t)}{R_{ij}(t) + \varepsilon}\right) * \left(x_{jd}(t) - x_{id}(t)\right) \quad (2)$$

$M_{aj}$ is the active gravitational mass related to agent $j$, $M_{pi}$ is the passive gravitational mass related to agent $i$ , $G(t)$ is a gravitational constant at time $t$, $\varepsilon$ is a small constant, and $R_{ij}(t)$ is the Euclidian distance between two agents $i$ and $j$ at time $t$. We consider the values of the two gravitational masses to be the same ($M_{pj} = M_{aj}$).

Gravitational and inertial masses are simply calculated by the fitness evaluation. A heavier mass means a fitter agent (corresponds to a lower value). This means that the better agents have higher attraction and walk more slowly. Assuming the equality of gravitational mass and inertial mass, values of masses are calculated using the value of fitness. The gravitational and inertial masses are updated by the following equations:

$$M_{ai} = M_{pi} = M_{ii} = M_i, i = 1,2, \dots, N \quad (3)$$

$M_{ii}$ is the inertial mass of the $i^{th}$ agent and $M_i$ is the overall mass of the $i^{th}$ agent.

$$m_i = \frac{fit_i(t) - worst(t)}{best(t) - worst(t)} \quad (4)$$

$$M_i = \frac{m_i(t)}{\sum_{j=1}^{N} m_j(t)} \quad (5)$$

$fit_i(t)$ represents the fitness value of agent $i$ at time $t$, and $worst(t)$ and $best(t)$ are defined as follows (for a minimization problem):

$$best(t) = \min_{j \in \{1,\dots,N\}} fit_j(t) \quad (6)$$

$$worst(t) = \max_{j \in \{1,\dots,N\}} fit_j(t) \quad (7)$$

It is to be noted that for a maximization problem, $max$ is used in place of $min$ and vice versa in Eqs. 6 and 7 respectively.

$G$ and $R_{ij}$ between two agents $i$ and $j$ are calculated as follows:

$$G(t) = G_o * exp(-\alpha * iter/maxiter) \quad (8)$$

$$R_{ij} = \sqrt[2]{\sum_{k=0}^{D} \left(X_{ik}(t) - X_{jk}(t)\right)^2} \quad (9)$$

α is the descending coefficient, $G_0$ indicates the initial gravitational constant, $iter$ is the current iteration, and $maxiter$ is the maximum number of iterations. In a problem space for the $d^{th}$ dimension, the total force that acts on agent $i$ is calculated by the following equation:

$$F_i^d(t) = \sum_{j=1, j \neq i}^{N} rand_j F_{ij}^d(t) \quad (10)$$

$rand_j$ is a random number generated with uniform distribution in the interval [0, 1]. The law of motion has also been utilized in this algorithm which states that acceleration of a mass is proportional to the resultant force and inverse of its mass, so the acceleration of all agents is calculated as follows:

$$a_{id}(t) = \frac{F_{id}(t)}{M_{ii}(t)} \quad (11)$$

$M_{ii}$ is the inertial mass of agent $i$. During optimization, the best-obtained solution so far is saved as $gbest$ following the concept of PSO. Eq. 12 was proposed as follows for combining PSO and GSA:

$$V_i(t+1) = rand * V_i(t) + c_1 * a_i(t) + c_2 * \left(gbest - X_i(t)\right) \quad (12)$$

$V_i(t)$is the velocity of agent $i$ at time $t$, $c_j$ is an accelerating factor, *rand* is a random number generated with a uniform distribution between 0 and 1, $a_i(t)$ is the acceleration of agent $i$ at time $t$, and $gbest$ is the best-obtained solution so far. In each iteration, the positions of agents are updated as follows:

$$X_i(t+1) = X_i(t) + V_i(t+1) \quad (13)$$

In PSOGSA, at first, all the agents are randomly initialized using uniform distribution. Each agent is considered as a candidate solution. After initialization, $F_{ij}^{d}$, $G(t)$, and $F_i^d(t)$ are calculated by Eqs. 2, 8 and 10 respectively. Whereas, the acceleration of particles are defined by Eq. 11. In each iteration, the best-attained solution should be updated. After calculating the acceleration and updating the best solution, the velocity of each agent is calculated by Eq. 12. Finally, the positions of agents are updated by Eq. 13. The process of updating velocities and positions is stopped when an end criterion is met.

### 2.2 Gravitational Particle Swarm (GPS)

GPS [26] is a swarm intelligence based hybrid algorithm which incorporates both the ideas of PSO and GSA. The particles under consideration in GPS update their respective positions based on both PSO and GSA velocities. Thus, GPS as a whole exploits both the social behavior of PSO as well as the population-based search pattern of GSA. Population and the respective particle positions and velocities are represented in the following manner.

$$X_i = (x_{i1}, \dots, x_{id}, \dots, x_{iD}), i = 1N \quad \text{where D is the dimension} \tag{14}$$

$$v_i^d(t+1)_{PSO} = w(t)v_i^d(t) + c_1 r_{i1}\left(pbest_i^d - x_i^d(t)\right) + c_2 r_{i2}\left(gbest_i^d - x_i^d(t)\right) \tag{15}$$

$$v_i^d(t+1)_{GSA} = rand_i * v_i^d(t) + F_i^d\,(t)/M_i^d\,(t) \tag{16}$$

$$v_i^d(t+1)_{GPS} = c_3 r_{i3} * v_i^d(t+1)_{PSO} + c_4(1 - r_{i3}) * v_i^d(t+1)_{GSA} \tag{17}$$

$$v_i^d(t+1) = x_i^d(t) + v_i^d(t+1)_{GPS} \tag{18}$$

Eq. 15 and Eq. 16 have been taken from PSO formulation [1] and GSA formulation [2] respectively, while Eq. 17 is the GPS velocity update based on Eq. 15 and Eq. 16. Of which, $r_{i3}$ is a random variable lying uniformly within [0, 1] to create stochastic impacts of PSO velocity and GSA velocity on GPS agent positions. $c_3$ and $c_4$ are two constants to determine the degree to which PSO and GSA velocities influence GPS. GPS is defined as GPS $(N, c_3, c_4)$. When both of $c_3$ and $c_4$ are valued at 1, GPS agents are stochastically impacted by equal influences of PSO and GSA.

### 2.3 Fuzzy Logic

Since inception in 1965 through the fuzzy set theory concept by Lofti Zadeh [37], it has been widely used and applied to a variety of fields like cancer classification [38], image segmentation [39], optimization [40] and so on. Most natural things cannot be defined by simple or convenient shapes or distributions. Fuzzy logic is the characterization of the truth value of a variable as a real number between [0,1]. Membership Functions (MFs) are used to define the fuzziness in a graphical form for eventual use in fuzzy set theory.

## 3. Proposed Model

The hybrid versions of PSO and GSA – PSOGSA and GPS suffer from premature convergence. This causes the algorithms to get trapped in a local optimum and it deters us to get the optimal solution. Hence, mutation has been applied to these hybrid versions of PSO and GSA, i.e. PSOGSA and GPS, to get rid of the problem of premature convergence. Due to the fast convergence nature of these hybrids, all the points tend to move fast towards the current best solution in each iteration. However, the point it converges to may not be the most optimal and to ensure that this does not occur, we propose a novel mutation approach called *centroid based fuzzy mutation*.

### 3.1 Fuzzy Logic-based Mutation

The concept of Fuzzy logic has widely been used for solving different research problems and it has applications from industry to academia. Following the concept of fuzzy logic, briefed in subsection 2.3, an MF gives a corresponding membership value of an operation. Any fuzzy set $F$ in the universal domain $U$ can be defined as a collection of ordered pairs. The mathematical representation of such a set is provided below:

$$F = \{(x, \mu_F(x)) | x \epsilon U\}$$

Where $\mu_F$ is the MF of $F$ with values in the range $[0,1]$ and $x$ is an element of information in universal set $U$. So, depending on the nature of the MF, an element of information can have different degrees of membership in the present domain. The elements having full membership form the core of the fuzzy set, the ones having non-zero membership are called support and the ones having non-zero but incomplete membership (< 1) are said to be the boundary of the fuzzy set.

**Formation of an MF:**

The MFs in Fuzzy logic has a crucial role in the overall performance of the fuzzy representation of the underlying problem. To be specific, the shape of a MF is important for a particular problem as it takes the decisive role of the fuzzy inference system. MFs can be of different shapes – Gaussian, triangular, trapezoidal etc. with the condition that the values of a MF vary from 0 to 1. An MF basically maps the given data with required degree of memberships. Deep understanding about the underlying problem can give us notion to know which shape of MF would fit the application under consideration.

There may be infinite number of ways to characterize fuzziness. The choice of which depends on the problem type. Therefore, apart from shape of a MF, deciding the interval as well as number of MFs is very important. For example, to model a control system in terms of temperature by fuzzy logic, it is vital to know how many MFs are required (e.g., high, medium, and low) along with the interval of membership values. These two parameters have a significant impact on the inference of a fuzzy logic-based system. Besides, observing the data distribution is another important factor. Many times, trial and error methods are applied for selecting the shape of MF shape as there is no exact method for selecting the MFs. The function may have an arbitrary curve, and it suits us in terms of efficiency, simplicity, and speed.

However, the number of MF has greater influence as it determines the computational time. Hence, the optimum model can be determined by varying the number/type of MFs for achieving best system performance. The work reported in [41] gives some idea about the shape which would be best if someone applies fuzzy logic as a universal approximator. In another work [42], a constrained interpolations concept was designed to fit a MF to a finite number of membership values. Some other works are found in the literature giving some directions of choosing MF [43]–[46]. The main concern is to break the 0–1 modelling, and it can be done by applying a triangular MF. Nevertheless, if the situation is more complex, we may require special type of MF. To make the best choice, a high-fidelity intuition based on adequate experience can give a satisfactory answer.

Using metaheuristic optimization methods and evolutionary optimization algorithms, fuzzy logic possesses the great flexibility toward its initial parameters regarding MFs [47]. Interested reader can find some useful information about MFs and some procedures (e.g., GA and neural network) to assign memberships to fuzzy variables [44].

This concept of fuzzy logic has been used for finding the probability of mutation being performed for a particle. At any moment, mutation is not completely certain or uncertain for a particle, instead, the membership value provides the probability of mutation. The proposed mutation-based model has been detailed in section 3.1.1. Thus, incorporation of fuzziness allows us to perform mutation in PSOGSA and GPS in a probabilistic manner.

### 3.2 Centroid based Fuzzy Mutation

This newly developed mutation helps the particles in the population to drift when there is the chance to pre-mature convergence. We have the following two important metrics when we consider a particle for mutation.

- The overall distance of the particles from the other particles
- History of the particles i.e. the change in the accuracies of the particles as a whole

First consideration checks the distance of a particle from the centroid of population. If the particles come closer to each other (overall distance among them decreases), then there is a possibility of premature convergence. To avoid this, we apply mutation to place the particles away from each other. This helps the particles to circumvent the local optimum and look for some other region in the search space. Instead of calculating the distance of a particle from every other particle, it is convenient to measure the distance of the particle from the centroid of all the particles. If this distance is less, it implies that the particle is close to the other particles and hence should be considered for mutation. So, we can see that the probability of mutation is inversely proportional to the distance of the particle from the centroid. In certain situations, it may so happen that a particular particle is residing at the centroid. This will make the distance to be 0 leading to infinite chances of mutation which is unacceptable. That is why we need to add 1 to the distance to ensure that this scenario never occurs. So, the contribution of distance to the probability of mutation is presented in Eq. 19 where $dist$ is the distance of the particle from the centroid and $P_d$ is the estimated contribution.

$$P_d = \frac{1}{1+dist} \tag{19}$$

Similar to distance, history of the particles may provide some important insight into the mutation probability. When the global best particle gets changed frequently over the iterations, it indicates that the particles are still exploring and trying to reach better solutions in the search space. But, on the other hand, if the global best particle is static over a significant number of iterations, it gives an indication that the particle might have gotten stuck in some local optima and is unable to explore different parts of the search space. In these scenarios, it becomes important to perform mutation on the particles to provide some perturbation to them, and thereby helping the particles to overcome the local optima and to reach the global optima. Thus, the probability of mutation should increase when the time for which the global best particles remains constant increases. We estimate the contribution of this historical information ($P_c$) to the probability of mutation using Eq. 20 where $unchanged$ is the number of iterations for which the global best particle has remained unaffected. We take $\alpha = 4$ and $\beta = 5$. Following these values, the probability of mutation increases as the value of $unchanged$ increases.

$$P_c = a + b * tanh\left(\left(\frac{unchanged}{\alpha}\right) - \beta\right) \tag{20}$$

where $a$ =0.5 and $b$ = 0.5. Depending on the value of the $unchanged$, the hyperbolic tan function will return a value in $[-1,1]$ which when multiplied by $b$ will be restricted in the range of $[-0.5,0.5]$. So, ultimately the value of $P_c$ is in [0,1]. We combine these two contributions using Eq. 21. Thus, if the $gbest$ is moving towards the optimal solution and the distance becomes large from the centroid then the point may explore an uncharted portion of the search space and hence the motion of the point is not disturbed. On the other hand, if the point goes closer to the centroid and $gbest$ remains unchanged for long duration then the point approaches to a well-explored portion of the search space, and therefore mutation is applied to disrupt its movement in order to explore a different region of the search space. The parameters $\rho$ and $\varphi$ assign weight to the contributions of distance and history in the probability equation. We have used $\rho$ as 0.6 and $\varphi$ as 0.4. Distance has been given more importance over history as there may be certain cases where although $gbest$ does not change, other particles change their positions. In this scenario, the particles do not get stuck but the approach will consider a probable convergence. So, to avoid that, we have assigned a lesser weight to history.

$$P_i = \rho * P_d + \varphi * P_c \tag{21}$$

$P_i$ denotes the probability of mutation for the $i^{th}$ particle of population. If $P_i$ is greater than a generated random value, the particle gets mutated, else no mutation takes place. In the proposed model, the term $P_i$ acts as membership value for fuzzy mutation. Mutation is not applicable for every particle. Instead $P_i$ helps us to find a set of fuzzy elements similar to the fuzzy set denoted as $F$ in section 2.3. After obtaining this set of fuzzy particles, if for particle $i$, $P_i$ is greater than a generated random value, the particle gets mutated, else no mutation takes place. We perform mutation using the following functions:

$$\Delta q = 0.5 * range * \left(\left(1 - \frac{count}{iter}\right)^2\right) \tag{22}$$

$$\Delta p = \min(\Delta q, P_{ij}) \tag{23}$$

where $\Delta p$ is the change in $j^{th}$ dimension's value of the particle, $range$ is the difference between the upper and lower limits of the domain of the benchmark function under consideration, $count$ denotes the current iteration number and $iter$ is the total number of iterations to be performed and $P_{ij}$ is the value of the $i^{th}$ particle in the $j^{th}$ dimension of the entire population. The value of $\Delta q$ gradually decreases over time to allow less disruption as the points converge. $\Delta p$ is restricted (in Eq. 23) to ensure the disruption in the motion of an agent is limited.

Mutation occurs alternatively subtracting or adding $\Delta p$ to $X_{ij}$ with the probability of addition and subtraction being half. This allows the point to move by a value of $\Delta p$ in any direction in the $D$ dimensional space. $\Delta p$ is evaluated D times as well as performed subtraction or addition of the value $\Delta p$ to each of the $X_{ij}$for all$j \in 1D$.

Trivially the importance of the fuzzy mutation is described in Figure 1 (A-D). Consider the scenario when there are two local minima – one having lesser value (desirable) than the other for a minimization problem. The particles are moving according to the motion defined by PSOGSA or GPS. There is a high chance that the particles will converge to a local minimum without even considering the other one. Figure 1A represents the force diagram of three particles M1, M2 and M3 when they are closer to local minima 1. The progression of the particles is shown in Figure 1B where they are almost converged to local minima 1. To avoid further convergence, the fuzzy mutation is used. Say only M1 and M2 pass the mutation criteria. The mutation direction of both particles are shown in Figure 1C. Depending on the extent of mutation, the particles may land into the final state described in Figure 1D where M1 and M2 have successfully avoided local minima 1 and moved towards local minima 2. We know that there are a lot of assumptions regarding this scenario but if we compare the GPS or PSOGSA with their mutated versions, the latter ones will always have better chances of avoiding such convergence problems.

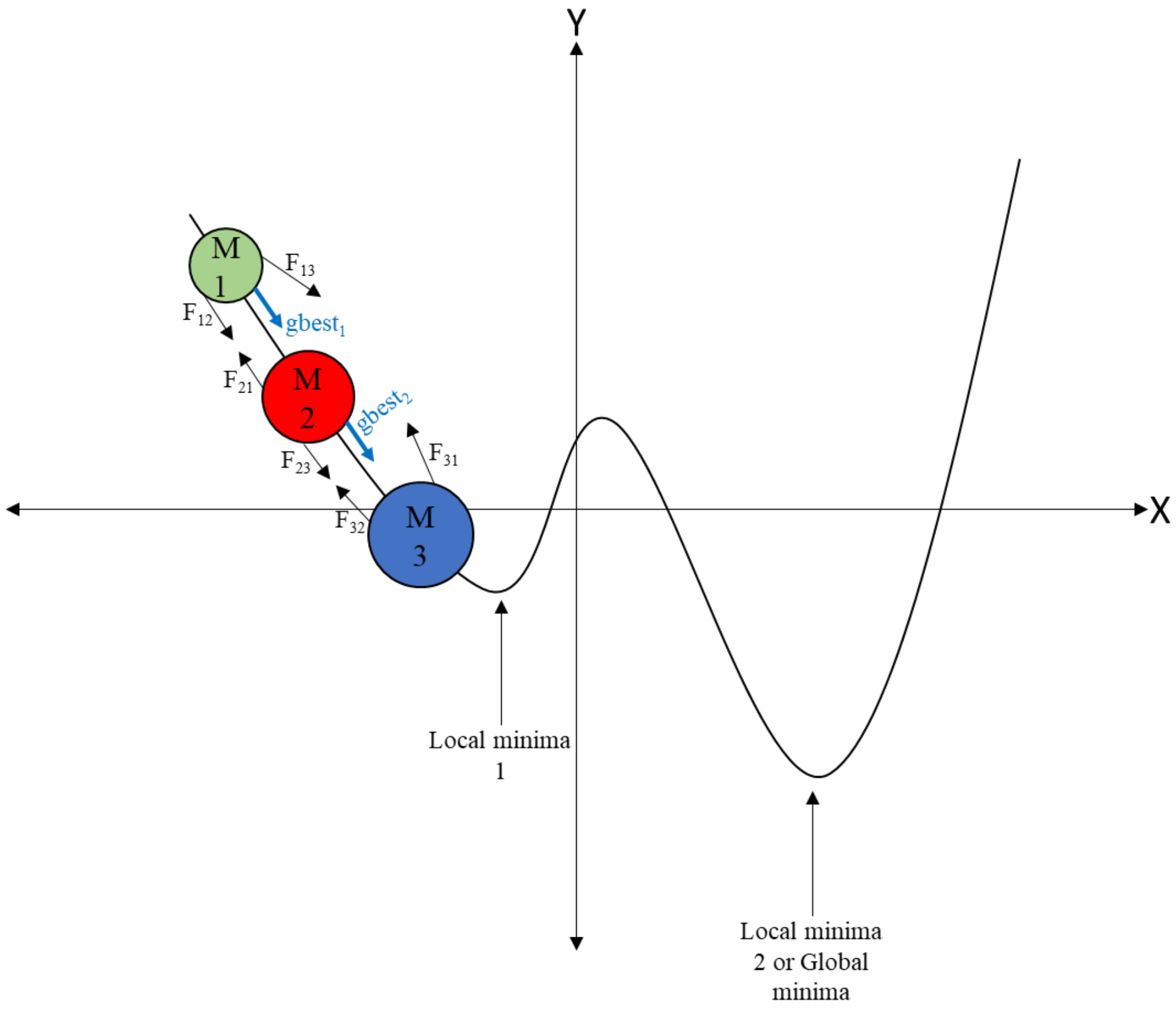

**A:** There are two minima in the graph - local minima 2 (maybe global minima) is having lesser value than local minima 1. Force diagram of 3 particles namely M1, M2 and M3 are shown when they are near local minima 1

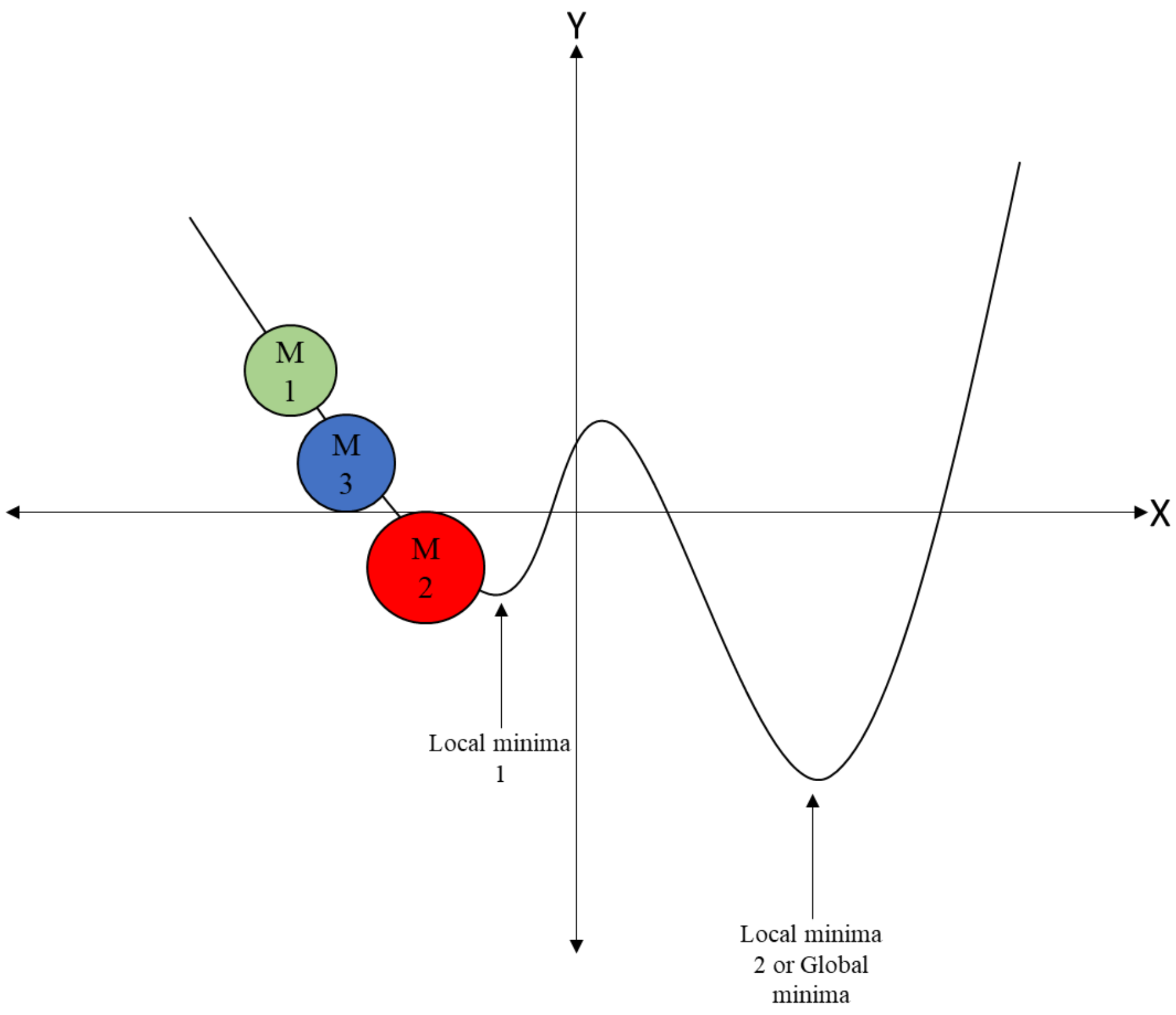

**B:** The particles are getting converged to the local minima 1 following the movement defined in PSOGSA (using Eqn. 1-13) or GPS (using Eqn. 14-18)

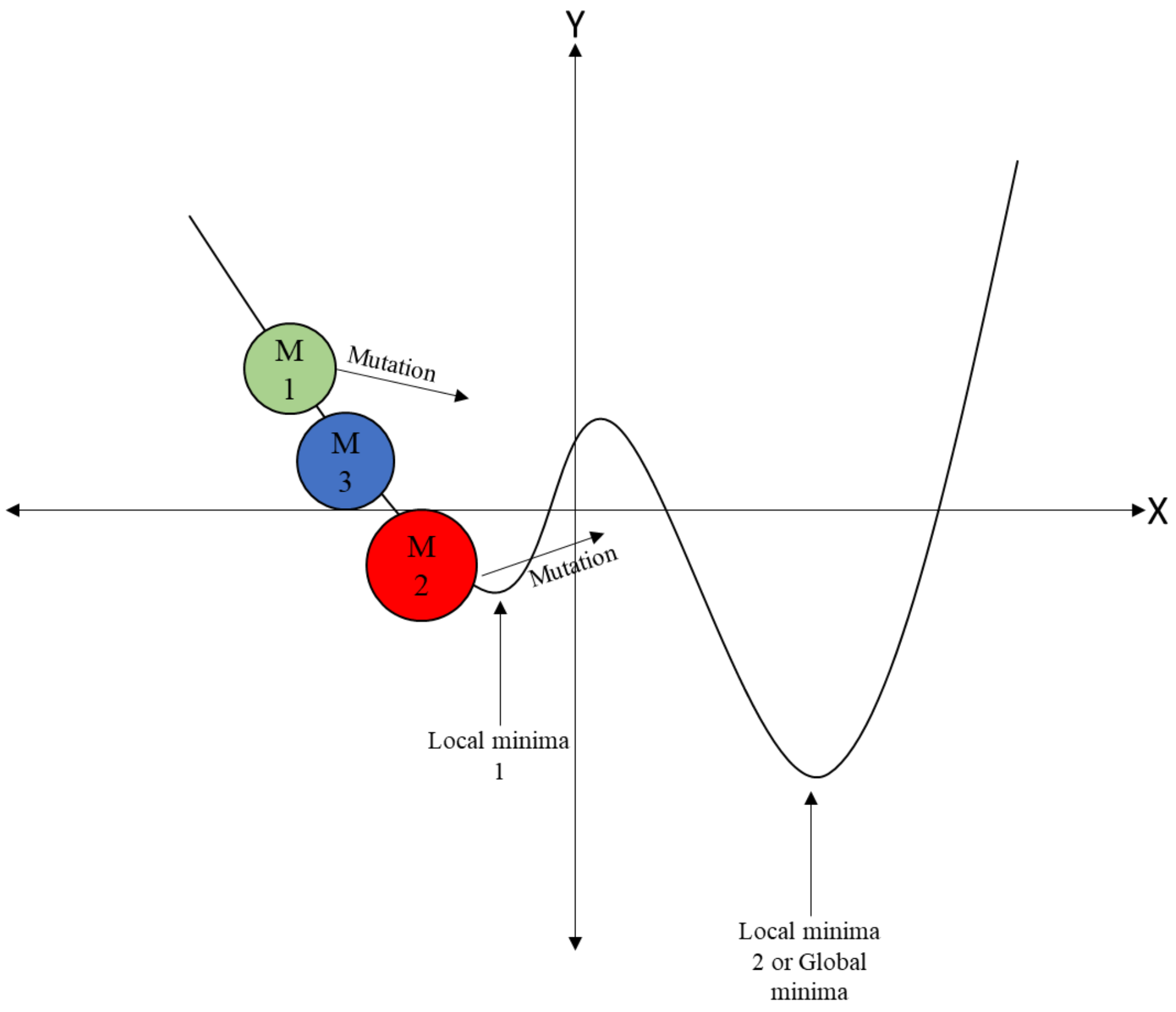

**C:** Depending on the probability value presented in Eqn. 21, say M1 and M2 are getting mutated while M3 is not. The extent of the mutation is calculated using Eqn. 23

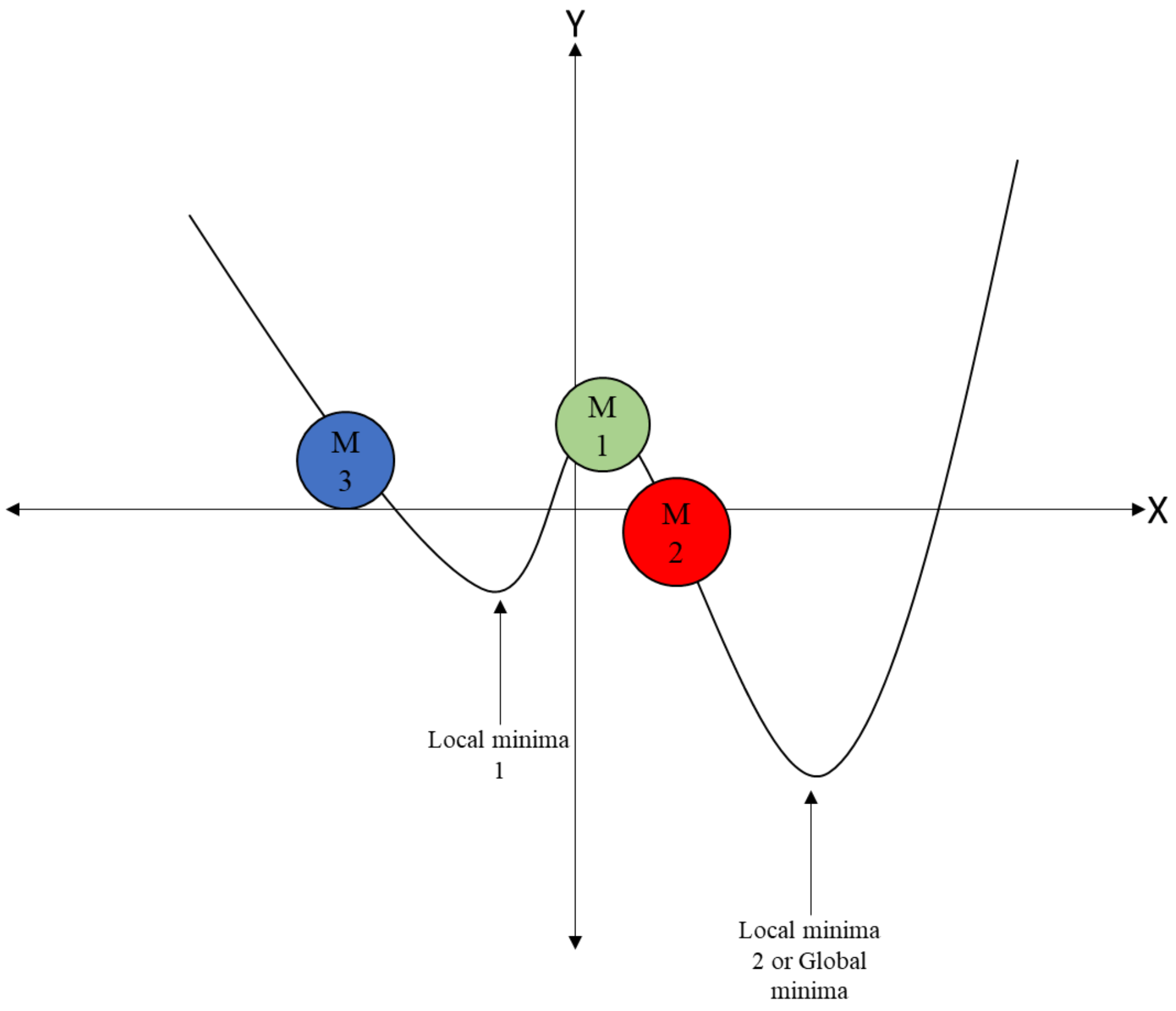

**D:** After mutation of the particles M1 and M2, they successfully circumvented local minima 1 and started moving towards local minima 2

**Figure 1(A-D):** Example illustrating the utility of mutation in PSOGSA and GPS to avoid convergence to local minima.

## 4. Experimental Results

The experiments in this work were performed on MATLAB in a PC having 4GB RAM. The proposed fuzzy mutation-based hybrid versions of PSOGSA and GPS, MPSOGSA and MGPS respectively, have been tested on several benchmark functions as given in Tables 1, 2 and 3 in the following section. The results are also given in section 4.1 and graphical depictions of convergence of the points are shown in section 4.2. The computation complexity of the fuzzy mutation algorithm is low and so our proposed algorithm has the same complexity as GPS and PSOGSA.

Three categories of functions used in Rashedi et al. [1] are adopted to test the MGPS and MPSOGSA. These categories include seven unimodal high-dimensional functions ($F_1 - F_7$ in Table 1); six multimodal high-dimensional functions ($F_8 - F_{13}$ in Table 2), and ten multimodal low-dimensional functions ($F_{13} - F_{23}$ in Table 3). Usually, the optimization of unimodal functions focuses on the convergence rate of a global optimum. Many local functional valleys exist due to which, finding the final optimum of the multimodal functions becomes difficult. While accuracy typically ranks as the most important consideration in optimization, no algorithm offers an absolute advantage in this regard. Obtaining a relatively good performance on optimization is one of the primary concerns in developing an optimization algorithm.

**Table 1**: Description of unimodal functions used in present work

| Function | Domain | Optimum | Position |
|---|---|---|---|
| $F_1 = \sum_{i=1}^{n} x_i^2$ | $[-100,100]^{30}$ | 0 | $(0)^{30}$ |
| $F_2 = \sum_{i=1}^{n} x_i \vee + \prod_{i=1}^{n} x_i \vee$ | $[-10,10]^{30}$ | 0 | $(0)^{30}$ |
| $F_3 = \sum_{i=1}^{n} \left( \sum_{j=1}^{i} x_j \right)^2$ | $[-100,100]^{30}$ | 0 | $(0)^{30}$ |
| $F_4 = max\{\lvert x_i \rvert, 1 \le i \le n\}$ | $[-100,100]^{30}$ | 0 | $(0)^{30}$ |
| $F_5 = \sum_{i=1}^{n-1} \left[ 100\left(x_{i+1} - x_i^2\right)^2 + (x_i - 1)^2 \right]$ | $[-30,30]^{30}$ | 0 | $(1)^{30}$ |
| $F_6 = \sum_{i=1}^{n} (x_i + 0.5)^2$ | $[-100,100]^{30}$ | 0 | $(0)^{30}$ |
| $F_7 = \sum_{i=1}^{n} i x_i^4 + random[0,1]$ | $[-1.28,1.28]^{30}$ | 0 | $(0)^{30}$ |

**Table 2**: Description of multimodal functions used in present work

| Function | Domain | Optimum | Position |
|---|---|---|---|
| $F_8 = \sum_{i=1}^{n} -x_i sin\left(\sqrt{\lvert x_i \rvert}\right)$ | $[-500,500]^{30}$ | -12569.5 | $(420.96)^{30}$ |
| $F_9 = \sum_{i=1}^{n} \left[x_i^2 - 10cos(2\pi x_i) + 10\right]$ | $[-5.12,5.12]^{30}$ | 0 | $(0)^{30}$ |
| $F_{10} = -20exp\left(-0.2\sqrt{\frac{1}{n}\sum_{i=1}^{n}x_i^2}\right) - exp\left(\frac{1}{n}\sum_{i=1}^{n}cos(2\pi x_i)\right) + 20 + e$ | $[-32,32]^{30}$ | 0 | $(0)^{30}$ |
| $F_{11} = \frac{1}{4000}\sum_{i=1}^{n}x_i^2 - \prod_{i=1}^{n}cos\left(\frac{x_i}{\sqrt{i}}\right) + 1$ | $[-600,600]^{30}$ | 0 | $(0)^{30}$ |
| $F_{12} = \frac{\pi}{n}\left\{10sin^2(\pi y_i) + \sum_{i=1}^{n-1}(y_i - 1)^2[1 + 10sin^2(\pi y_{i+1})] + (y_n - 1)^2\right\} + \sum_{i=1}^{n}u(x_i, 10,100,4)$ <br> $y_i = 1 + \frac{x_i + 1}{4}$ <br> $u(x_i, a, k, m) = \begin{cases} k(x_i - a)^m, x_i > a \\ 0, -a \le x \le a \\ k(-x_i - a)^m, x_i \leftarrow a \end{cases}$ | $[-50,50]^{30}$ | 0 | $(1)^{30}$ |
| $F_{13} = 0.1\left\{sin^2(3\pi x_1) + \sum_{i=1}^{n}(x_i - 1)^2[1 + sin^2(3\pi x_i)] + (x_n - 1)^2[1 + sin^2(2\pi x_n)]\right\} + \sum_{i=1}^{n}u(x_i, 5,100,4)$ | $[-50,50]^{30}$ | 0 | $(1)^{30}$ |

**Note:** *e* is Euler's constant

**Table 3**: Multimodal functions with fixed dimension used in present work

| Function | Domain | Optimum | Position |
|---|---|---|---|
| $F_{14} = \left(\frac{1}{500} + \sum_{j=1}^{25}\frac{1}{j + \sum_{i=1}^{2}(x_i - a_{ij})^6}\right)^{-1}$ | $[-65.53,65.53]^2$ | 1 | (-32,32) |
| $F_{15} = \sum_{i=1}^{11}\left[a_i - \frac{x_1(b_1^2 + b_1x_2)}{b_1^2 + b_1x_3 + x_4}\right]^2$ | $[-5,5]^4$ | 0.00030 | (0.1928,0.1908,0.1231, 0.1358) |
| $F_{16} = 4x_1^2 - 2.1x_1^4 + \frac{1}{3}x_1^6 + x_1x_2 - 4x_2^2 + 4x_2^4$ | $[-5,5]^2$ | -1.0316 | (0.089,0.712), (-0.089,0.712), |
| $F_{17} = \left(x_2 - \frac{5.1}{4\pi^2}x_1^2 + \frac{5}{\pi}x_1 - 6\right)^2 + 10\left(1 - \frac{1}{8\pi}\right)cosx_1 + 10$ | $[-5,10] \times [0,15]$ | 0.398 | (-3.14,12.27), (3.14,12.275), (9.42,2.42) |
| $F_{18} = [1 + (x_1 + x_2 + 1)^2(19 - 14x_1 + 3x_1^2 - 14x_2 + 6x_1x_2 + 3x_2^2)] \times [30 + (2x_1 - 3x_2)^2 \times (18 - 32x_1 + 12x_1^2 + 48x_2 - 36x_1x_2 + 27x_2^2)]$ | $[-5,5]^2$ | 3 | [0, -1] |
| $F_{19} = -\sum_{i=1}^{4} c_i exp\left(-\sum_{j=1}^{3} a_{ij}\left(x_j - p_{ij}\right)^2\right)$ | $[0,1]^3$ | -3.86 | (0.114, 0.556, 0.852) |
| $F_{20} = -\sum_{i=1}^{4} c_i exp\left(-\sum_{j=1}^{6} a_{ij}\left(x_j - p_{ij}\right)^2\right)$ | $[0,1]^6$ | -3.32 | (0.201, 0.15, 0.477, 0.275, 0.311, 0.657) |
| $F_{21} = -\sum_{i=1}^{5}[(X - a_i)(X - a_i)^T + c_i]^{-1}$ | $[0,10]^4$ | -10.1532 | $5a_{ij}$ |
| $F_{22} = -\sum_{i=1}^{7}[(X - a_i)(X - a_i)^T + c_i]^{-1}$ | $[0,10]^4$ | -10.4028 | $7a_{ij}$ |
| $F_{23} = -\sum_{i=1}^{10}[(X - a_i)(X - a_i)^T + c_i]^{-1}$ | $[0,10]^4$ | -10.5363 | $10a_{ij}$ |

**Note**: For detailed description of the functions of Table 3 refer to Appendix A of Rashedi [1].

### 4.1 Results on Benchmark Functions

This section illustrates the results obtained by the proposed algorithms over 23 benchmark functions from unimodal, multi-modal and multi-modal with fixed dimension categories. The final results are also compared with some state-of-the-art algorithms to justify the applicability of the proposed mutation models.

### 4.1.2 Parameter Tuning

In order to obtain proper results of the proposed models, some experimentations have been performed to fine-tune the parameters present in the algorithms. There are mainly four parameters - $\alpha$, $\beta$ as given in Eqn. 20 and $\rho$, $\varphi$ as given in Eqn. 21. Apart from these four parameters, there are number of iterations and population size which should have a fixed value for a uniform testing environment. For all the experimentations and comparison, we have fixed the population size to be 50 and used 500 iterations for $F_1$-$F_7$, 1000 iterations for $F_8$-$F_{23}$. For the previously mentioned four parameters, the qualities of the solutions are checked by varying their values and finally the most optimal combination out of them is selected.

In order to select the optimal values of the parameters, one function from each category of the benchmark functions have been selected – $F_1$ from the set of unimodal functions, $F_{10}$ from the set of multi-modal functions, and $F_{15}$ from the set of multi-modal functions with fixed dimension category. The testing is done for MGPS algorithm. At first, $\alpha$, $\beta$ values of the model have been varied and tested on these three functions followed by the testing of $\rho$, $\varphi$ values over the same three functions. Graphical representations of the results obtained through the testing are provided in Figure 2.

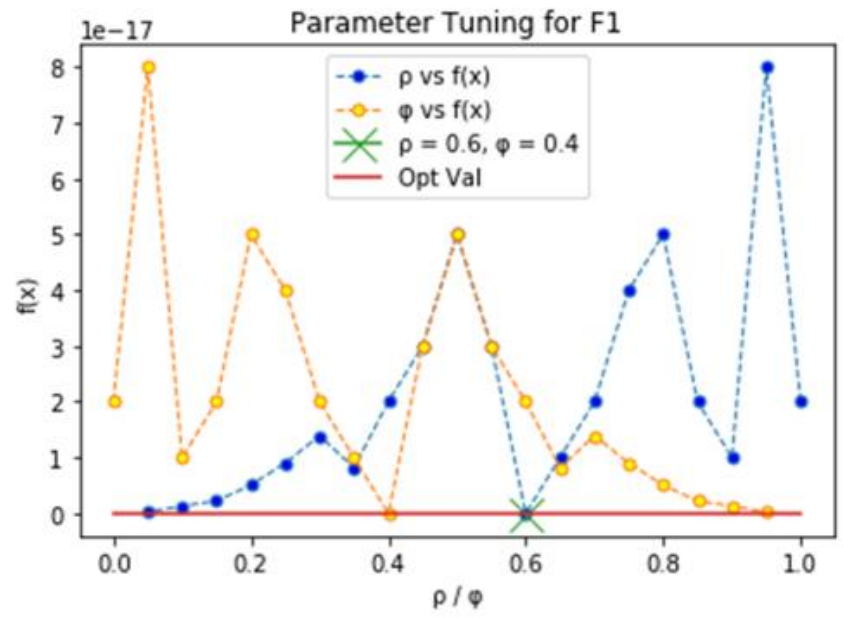

A: Graph showing the function value plotted against the parameters ρ/ϕ for $F_1$

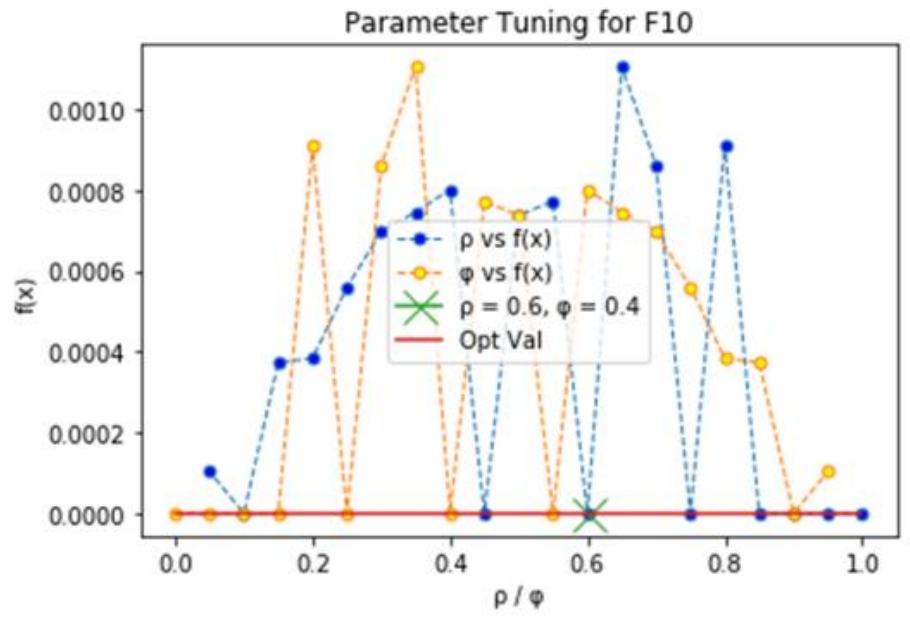

B: Graph showing the function value plotted against the parameters ρ/ϕ for $F_{10}$

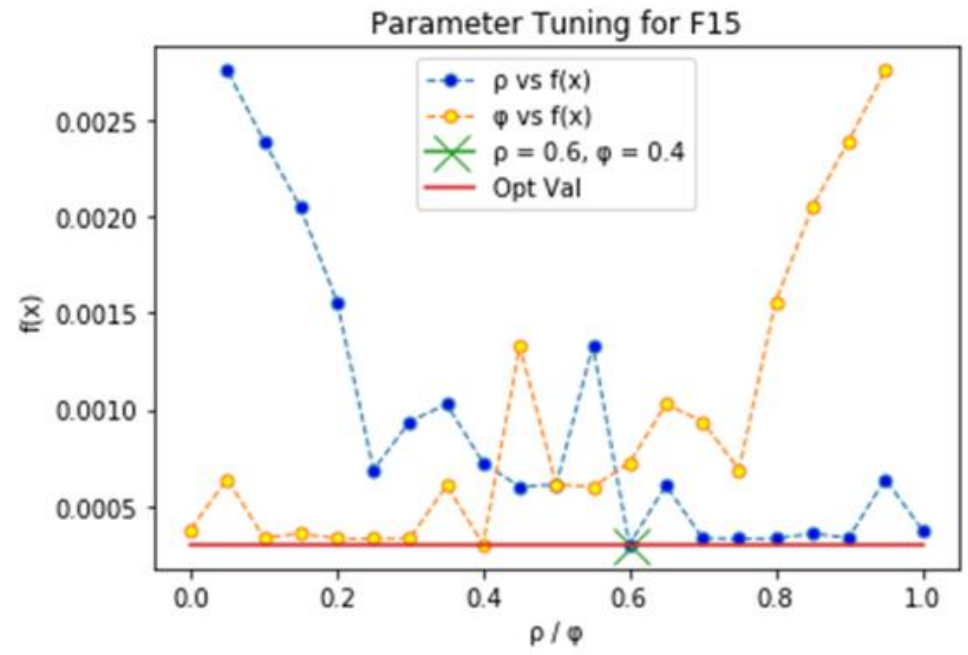

C: Graph showing the function value plotted against the parameters ρ/ϕ for $F_{15}$

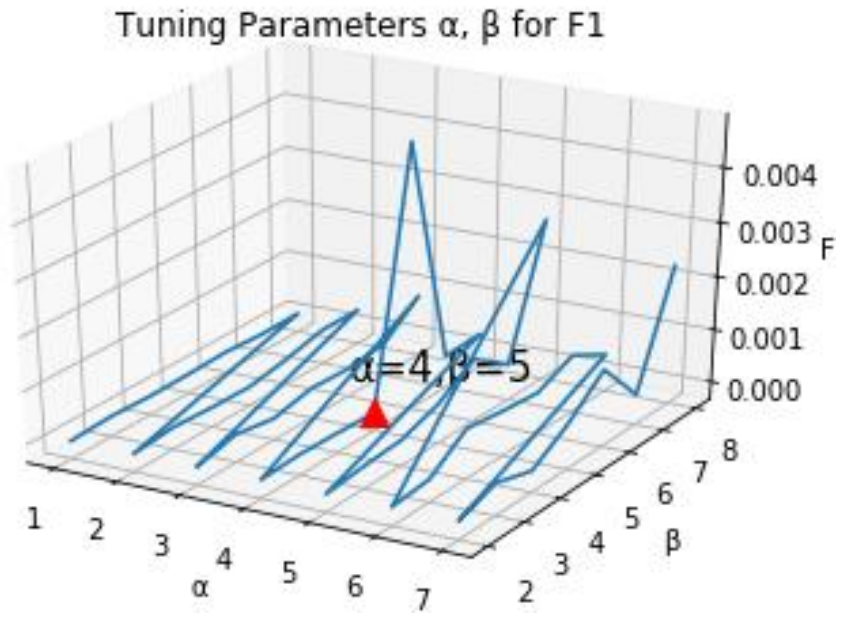

D: Graph showing the function value plotted against α, β the parameters ρ/ϕ for $F_1$

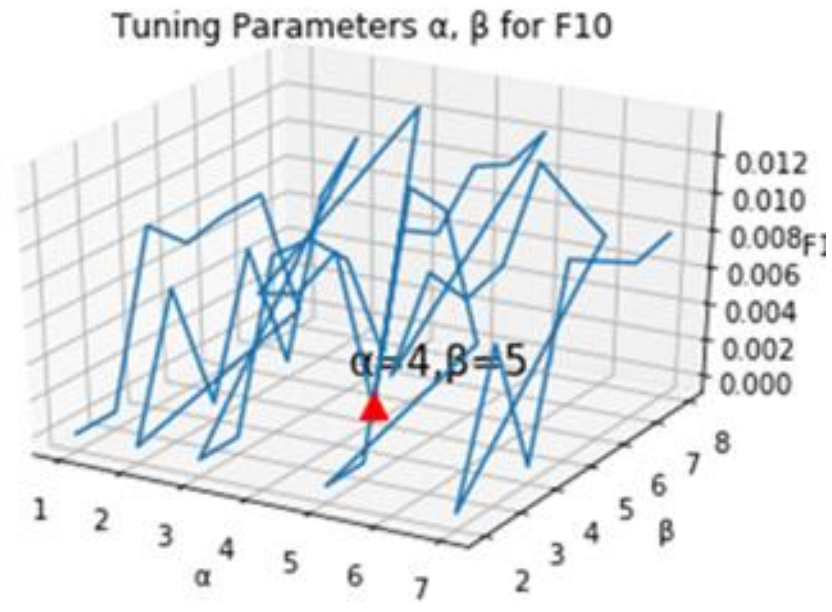

E: Graph showing the function value plotted against α, β the parameters ρ/ϕ for $F_{10}$

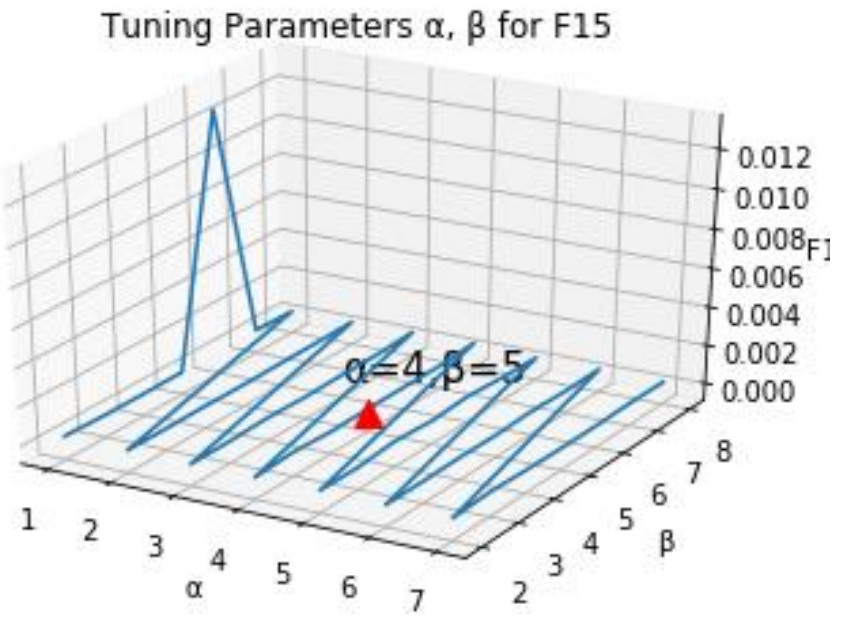

F: Graph showing the function value plotted against α, β the parameters ρ/ϕ for $F_{15}$

Figure 2 (A-C): Graphical representations of the results obtained via varying ρ/ϕ for F1, $F_9$, $F_{15}$.

Figure 2 (D-F): Graphical representations of the results obtained via varying α, β for F1, $F_9$, $F_{15}$.

After testing, the final values for the parameters are selected as mentioned in Table 4. For rest of the experimentations, these values have been used.

**Table 4:** Values of different parameters used in the proposed models.

| Parameter | Value |
|---|---|
| Population size | 50 |
| No. of iterations | 500 iterations for $F_1$-$F_7$, 1000 iterations for $F_8$-$F_{23}$ |
| $\alpha$ | 4 |
| $\beta$ | 5 |
| $\rho$ | 0.6 |
| $\varphi$ | 0.4 |

### 4.1.3 Comparison with State-of-the-art

The results of our two optimization models have been tabulated in Table 5 against four other optimization approaches namely, PSO, GSA, PSOGSA and GPS. PSO is used to represent PSO (50); the value 50 signifies the population size, that is, the number of population points. GSA represents GSA (50); GPS represents GPS (50, 1.0, 1.0) with equal influences from PSO and GSA. PSOGSA as well as our proposed models, MGPS and MPSOGSA follow the same parameters. All the used algorithms have been run independently for 25 times out of which only the best 20 results have been used to measure the average, best, worst and standard deviation (Sd.) values. The minimum most value obtained for the corresponding fitness function is taken to be the better result here. The performance of each algorithm has been determined by comparing the average fitness value given by the corresponding algorithm for each function and then, if need be, the best (minimum) value returned by an algorithm for that function and then the standard deviation (Sd.) value (if the average and best values are same). The most optimum value has been bolded and underlined. The second-best value has been bolded. MGPS outperforms GPS 13 out of 23 times (56.52%) and MPSOGSA outperforms PSOGSA 17 times out of 23 (73.91%).

We have further used the results of contemporary optimization algorithms including SSO (Social Spider Optimization, 2013) [30], SLC (Soccer League Championship Algorithm, 2014) [32], SCA (Sine Cosine Algorithm, 2016) [28], OBSCA (Opposition-based SCA, 2017) [29] and VPL (Volleyball Premier League Algorithm, 2018) [33] for comparison. The fitness values for the algorithms SCA (Sine Cosine Algorithm), OBSCA (Opposition-based SCA), OBPSO (Opposition-based PSO) and SSO (Social Spider Optimization) have been taken from the paper on OBSCA(2017) [29] to construct tables 8, 9 and 10 and those of tables 5, 6 and 7 for LCA (League Championship Algorithm) and SLC (Soccer League Competition Algorithm) have been taken from the paper on VPL(2018) [33]. Our algorithms collectively outperform OBPSO in over 85% cases, SCA in over 85% cases, SSO in nearly 70% cases, OBSCA in 70% cases. The values for the algorithms VPL, LCA, SLC have been referred from the paper on VPL (2018) [33]. It can be seen that our algorithms collectively outperform SLC in over 75% cases, LCA in nearly 60% cases and VPL in 56.52% cases.

The results in Tables 5-8 show that the proposed algorithms perform well in the category of multi-modal functions. Out of 16 functions, the proposed algorithms outperform the others in 8 functions, which show that the use of mutation has allowed the algorithms to avoid being stuck in local optima. It should be noted that in case of unimodal functions as well the proposed algorithms perform quite well in comparison to their parents GPS and GSAPSO. In total the proposed algorithms are the best in 9 cases and the second best in 4 cases. In comparison VPL it is best in 6 cases and second best in 4 cases. LCA on the other hand is best in 5 cases and second best in 2 cases. This shows that rank wise the proposed algorithms perform quite well.

In comparison with OBSCA, SCS, OBPSO and SSO in terms of only best, average and standard deviation, it can be seen that proposed algorithms have performed quite well as well.

The proposed algorithms are best in 10 cases and second best in 12 cases. So, in all, except 2 functions the proposed algorithms have rank of 1 or 2.

The proposed algorithms are very capable of avoiding local minima and thus perform quite well for the category of multi-modal functions. The mutation though in some cases causes fluctuation as seen in section 4.2 as mutation causes the particles to move in all directions which causes both an increase and decrease in fitness value. The main applicability of the solution is for problems whose fitness function corresponds to multi-modal functions of fixed dimension. The  No Free lunch theorem [49] points out that the fact that no one algorithm can outperform all others in all cases and this is what keeps research alive in this field. However, the better performance in comparison to other algorithms in most cases shows the applicability and effectiveness of the proposed algorithms.

**Table 5**: Comparison of results of optimization algorithms on unimodal functions. The best, average and standard deviation of different algorithms have been given.

| Function | Value Heads | GSA | PSO | GPS | PSOGSA | LCA | VPL | SLC | **Proposed Algorithms** | |
|---|---|---|---|---|---|---|---|---|---|---|
| | | | | | | | | | MGPS | MPSOGSA |
| $F_1$ | Best<br>Avg.<br>Sd. | 1.1E-17<br>2.0E-17<br>5.5E-18 | 1.1E-15<br>1.3E-11<br>8.8E-11 | 6.6E-19<br>1.2E-18<br>3.0E-19 | 3.29E-19<br>4.74E-19<br>8.04E-19 | 1.41E-48<br>3.25E-46<br>1.79E-46 | **0.00E+00**<br>**7.81E-132**<br>**4.20E-131** | **1.90E-166**<br>**4.40E-160**<br>**3.91E-80** | 4.12E-21<br>9.38E-19<br>1.57E-18 | 3.10E-13<br>1.92E-09<br>3.67E-09 |
| $F_2$ | Best<br>Avg.<br>Sd. | 1.4E-08<br>2.4E-08<br>4.4E-09 | 4.4E-09<br>2.9E-06<br>1.3E-05 | 3.3E-09<br>5.2E-09<br>9.0E-10 | 2.47E-09<br>2.93E-09<br>2.64E-10 | **4.58E-25**<br>**9.79E-25**<br>**1.49E-24** | **1.12E-102**<br>**1.13E-90**<br>**5.13E-90** | 1.12E-125<br>8.85E-06<br>9.84E+03 | 2.38E-22<br>1.40E-20<br>2.21E-20 | 2.11E-14<br>1.09E-10<br>4.48E-10 |
| $F_3$ | Best<br>Avg.<br>Sd. | 7.5E+01<br>2.3E+02<br>1.0E+02 | 1.9E+01<br>1.2E+02<br>7.5E+01 | 3.1E+00<br>9.7E+01<br>1.1E+02 | 2.92E+02<br>1.82E+03<br>4.82E+02 | 3.26E+03<br>1.12E+03<br>6.06E+03 | **1.93E-33**<br>**8.16E-04**<br>**2.85E-03** | 2.58E-25<br>2.11E-02<br>1.33E+01 | **1.17E-06**<br>**5.57E-07**<br>**5.52E-06** | 2.5E-02<br>9.98E-01<br>7.1E+01 |
| $F_4$ | Best<br>Avg.<br>Sd. | 2.1E-09<br>6.4E-02<br>2.5E-01 | 1.4E-01<br>4.2E-01<br>1.9E-01 | 8.2E-10<br>1.3E+00<br>9.8E-01 | 1.3E+01<br>2.2E+01<br>3.89E+00 | 1.49E+00<br>5.09E-01<br>2.63E+00 | **0.00E+00**<br>**1.54E-29**<br>**3.96E-29** | 8.96E-30<br>3.07E-04<br>1.03E+00 | **1.17E-06**<br>**5.52E-06**<br>**1.88E-06** | 8.30E-05<br>3.19E-04<br>2.74E-04 |
| $F_5$ | Best<br>Avg.<br>Sd. | 2.6E+01<br>2.8E+01<br>1.0E+01 | 2.5E+01<br>2.7E+01<br>8.4E+00 | 2.3E+01<br>2.6E+01<br>8.8E+00 | 1.6E+01<br>2.6E+01<br>2.5E+00 | **9.86E-03**<br>**8.42E-01**<br>**7.15E-01** | 2.58E+01<br>2.62E+01<br>2.76E-01 | 3.96E+01<br>3.00E+01<br>3.32E-01 | **4.19E+00**<br>**2.45E+01**<br>**9.47E+00** | 2.24E+01<br>2.51E+01<br>6.65E-01 |
| $F_6$ | Best<br>Avg.<br>Sd. | 7.4E-18<br>1.9E-17<br>6.4E-18 | 8.3E-16<br>1.3E-12<br>7.1E-12 | 6.0E-19<br>1.2E-18<br>3.3E-19 | **3.30E-19**<br>**5.05E-19**<br>**9.40E-20** | **0.00E+00**<br>**0.00E+00**<br>**0.00E+00** | 1.82E-05<br>4.09E-04<br>5.33E-04 | 1.02E+01<br>1.05E+01<br>1.12E-01 | 4.79E-06<br>2.21E-02<br>2.39E-02 | 1.75E-05<br>1.26E-02<br>5.59E-02 |
| $F_7$ | Best<br>Avg.<br>Sd. | 8.4E-03<br>2.8E-02<br>1.7E-02 | 1.7E-03<br>7.0E-03<br>2.5E-03 | 1.1E-03<br>3.1E-03<br>1.2E-03 | 1.5E-02<br>3.3E-02<br>9.3E-03 | 1.56E-02<br>9.48E-03<br>3.44E-03 | **4.67E-05**<br>**1.93E-03**<br>**1.36E-03** | **3.43E-01**<br>**3.76E-06**<br>**1.60E+01** | 8.51E-04<br>1.73E-02<br>1.01E-02 | 1.13E-04<br>1.6E-02<br>1.0E-02 |

**Table 6**: Comparison of results of optimization algorithms on multimodal functions. The best, average and standard deviation of different algorithms have been given.

| Function | Value Heads | GSA | PSO | GPS | PSOGSA | LCA | VPL | SLC | **Proposed Algorithms** | |
|---|---|---|---|---|---|---|---|---|---|---|
| | | | | | | | | | MGPS | MPSOGSA |
| $F_8$ | Best<br>Avg.<br>Sd. | -4.2E+03<br>-2.7E+03<br>4.7E+02 | **-1.0E+04**<br>**-9.0E+03**<br>**5.2E+02** | -8.9E+03<br>-7.5E+03<br>7.7E+02 | **-8.85E+03**<br>**-8.09E+03**<br>**4.71E+02** | -3.72E+03<br>-7.53E+02<br>2.21E+03 | -1.19E+112<br>-4.68E+90<br>2.15E+111 | -1.33E-25<br>-7.07E+03<br>1.24E+02 | -9.11E+03<br>-7.46E+03<br>6.21E+02 | -7.86E+03<br>-7.31E+03<br>4.76E+02 |
| $F_9$ | Best<br>Avg.<br>Sd. | 9.0E+00<br>1.7E+01<br>4.3E+00 | 1.8E+01<br>4.1E+01<br>1.5E+01 | 9.0E+00<br>2.1E+01<br>6.1E+00 | 4.48E+01<br>7.42E+01<br>1.1E+01 | **0.00E+00**<br>**0.00E+00**<br>**0.00E+00** | **0.00E+00**<br>**0.00E+00**<br>**0.00E+00** | **0.00E+00**<br>**0.00E+00**<br>**0.00E+00** | 1.14E-13<br>3.47E+00<br>2.84E+00 | 1.29E-11<br>2.89E-07<br>1.06E-06 |
| $F_{10}$ | Best<br>Avg.<br>Sd. | 2.2E-09<br>3.4E-09<br>4.1E-10 | 4.6E-09<br>9.1E-08<br>2.0E-07 | 5.4E-10<br>8.8E-10<br>1.3E-10 | 4.32E-10<br>5.07E-10<br>4.70E-11 | **2.22E-14**<br>**4.93E-15**<br>**3.74E-14** | **8.88E-16**<br>**8.88E-16**<br>**9.86E-32** | 7.08E-16<br>7.05E-14<br>1.86E-29 | 2.86E-11<br>1.42E-08<br>3.59E-08 | 5.62E-07<br>5.83E-05<br>5.28E-05 |
| $F_{11}$ | Best<br>Avg.<br>Sd. | 2.0E+00<br>4.3E+00<br>1.6E+00 | 5.1E-15<br>1.2E-02<br>1.2E-02 | 0.0E+00<br>2.3E-02<br>3.0E-02 | 2.86E-06<br>2.33E-01<br>3.5E-01 | 1.28E-13<br>5.84E-03<br>2.65E-03 | **0.00E+00**<br>**0.00E+00**<br>**0.00E+00** | **0.00E+00**<br>**0.00E+00**<br>**0.00E+00** | 0.0E+00<br>8.49E-05<br>2.96E-04 | 6.79E-14<br>1.09E-11<br>1.75E-11 |
| $F_{12}$ | Best<br>Avg.<br>Sd. | 6.2E-20<br>2.5E-02<br>6.1E-02 | 1.6E-18<br>1.5E-02<br>3.6E-02 | 4.7E-21<br>5.0E-02<br>1.3E-01 | 1.01E+00<br>4.46E+00<br>1.80E+00 | **1.57E-32**<br>**1.09E-47**<br>**1.57E-32** | **1.11E-06**<br>**2.58E-05**<br>**1.74E-05** | 0.00E+00<br>1.04E+01<br>8.46E-82 | 1.00E-03<br>1.9E-03<br>5.32E-04 | 2.66E-08<br>1.2E-02<br>2.4E-03 |
| $F_{13}$ | Best<br>Avg.<br>Sd. | 1.22E-18<br>2.1E-18<br>5.0E-19 | **9.9E-131**<br>**2.0E-31**<br>**4.3E-31** | 3.75E-08<br>8.48E-02<br>8.0E-02 | 9.29E-20<br>2.2E-03<br>4.5E-03 | **1.35E-32**<br>**5.47E-48**<br>**1.35E-32** | 2.63E-05<br>4.18E-04<br>4.84E-04 | 7.61E-01<br>1.00E+00<br>3.09E-01 | 2.56E-02<br>4.33E-01<br>3.1E-01 | 5.93E-01<br>9.73E-01<br>1.88E-01 |

**Table 7**: Comparison of results of optimization algorithms on multi-modal functions of fixed dimension. The best, average and standard deviation of different algorithms have been given.

| Function | Value Heads | GSA | PSO | GPS | PSOGSA | LCA | VPL | SLC | **Proposed Algorithms** | |
|---|---|---|---|---|---|---|---|---|---|---|
| | | | | | | | | | MGPS | MPSOGSA |
| $F_{14}$ | Best | 1.0E+00 | **1.0E+00** | 1.0E+00 | 1.0E+00 | 9.98E-01 | 9.98E-01 | 1.06E-04 | **1.0E+00** | 1.0E+00 |
| | Avg. | 3.8E+00 | **1.0E+00** | 1.0E+00 | 1.39E+00 | 3.33E-16 | 9.98E-01 | 0.00E+00 | **1.0E+00** | 1.51E+00 |
| | Sd. | 2.6E+00 | **3.2E-17** | 5.8E-01 | 8.75E-01 | 9.98E-01 | 2.32E-13 | 9.86E-32 | **2.84E-17** | 6.81E-01 |
| $F_{15}$ | Best | 1.4E-03 | 3.1E-04 | **3.1E-04** | 3.1E-04 | 9.95E-04 | 2.45E-05 | 0.00E+00 | 3.53E-04 | **3.23E-04** |
| | Avg. | 4.1E-03 | 1.2E-03 | **4.1E-04** | 7.16E-04 | 4.84E-04 | 1.25E-03 | 2.22E-03 | 9.05E-04 | **3.59E-04** |
| | Sd. | 3.2E-03 | 4.0E-03 | **3.4E-04** | 2.19E-04 | 1.29E-03 | 3.08E-04 | 6.66E-16 | 4.43E-14 | **4.07E-05** |
| $F_{16}$ | Best | -1.0E+00 | -1.0E+00 | -1.0E+00 | -1.3E+00 | -1.01E+00 | -1.03E+00 | 2.92E-04 | **-1.03E+00** | **-1.03E+00** |
| | Avg. | -1.0E+00 | -1.0E+00 | -1.0E+00 | -1.3E+00 | 3.26E-01 | -1.03E+00 | 4.49E-04 | **-1.03E+00** | **-1.03E+00** |
| | Sd. | 4.0E-16 | 2.3E-16 | 2.8E-16 | 2.28E-16 | 4.50E-01 | 2.56E-06 | 1.14E-06 | **0.00E+00** | **6.83E-17** |
| $F_{17}$ | Best | 4.0E-01 | 4.0E-01 | 4.0E-01 | **3.98E-01** | 3.98E-01 | 3.98E-01 | 0.00E+00 | 3.98E-01 | **3.98E-01** |
| | Avg. | 4.0E-01 | 4.0E-01 | 4.0E-01 | **3.98E-01** | 3.98E-01 | 3.98E-01 | 1.78E-15 | 3.98E-01 | **3.98E-01** |
| | Sd. | 3.4E-16 | 3.4E-16 | 3.4E-16 | **0.00E+00** | 1.11E-16 | 2.69E-06 | 0.00E+00 | 1.49E-05 | **0.00E+00** |
| $F_{18}$ | Best | 3.0E+00 | 3.0E+00 | 3.0E+00 | **3.00E+00** | 3.00E+00 | 3.00E+00 | 1.93E-14 | **3.00E+00** | 3.00E+00 |
| | Avg. | 3.0E+00 | 3.0E+00 | 3.0E+00 | **3.00E+00** | 3.00E+00 | 3.00E+00 | 4.81E-82 | **3.00E+00** | 3.00E+00 |
| | Sd. | 2.2E-15 | 3.1E-15 | 1.6E-15 | **8.40E-16** | 9.33E-07 | 7.58E-05 | 2.8E-152 | **6.09E-16** | 9.38E-15 |
| $F_{19}$ | Best | -3.9E+00 | -3.9E+00 | -3.9E+00 | **-3.86E+00** | -1.96E-01 | -3.85E+00 | -4.10E-77 | **-3.85E+00** | -3.88E+00 |
| | Avg. | -3.6E+00 | -3.9E+00 | -3.9E+00 | **-3.86E+00** | 4.96E-02 | -3.77E+00 | -1.36E+01 | **-3.87E+00** | -3.88E+00 |
| | Sd. | 3.0E-01 | 3.1E-15 | 3.1E-15 | **2.19E-15** | 2.81E-02 | 9.37EE-02 | 1.27E-04 | **2.70E-02** | 1.63E-16 |
| $F_{20}$ | Best | -3.3E+00 | -3.3E+00 | -3.31E+00 | **-3.32E+00** | -3.00E+00 | -3.32E+00 | -8.60E+03 | -3.32E+00 | **-3.32E+00** |
| | Avg. | -1.9E+00 | -3.3E+00 | -3.3E+00 | **-3.31E+00** | 5.89E-01 | -3.28E+00 | 8.88E-16 | -3.26E+00 | **-3.32E+00** |
| | Sd. | 5.4E-01 | 5.5E-02 | 2.4E-02 | **6.07E-01** | 1.54E+00 | 5.41E-02 | 0.00E+00 | 4.75E-02 | **0.00E+00** |
| $F_{21}$ | Best | -5.1E+00 | -1.0E+01 | **-1.0E+01** | -1.02E+01 | -3.40E+00 | **-1.02E+01** | 3.00E-25 | -4.99E+00 | -1.02E+01 |
| | Avg. | -5.1E+00 | -7.2E+00 | **-8.5E+00** | -6.17E+00 | 5.67E-01 | **-9.30E+00** | 9.97E-01 | -3.61+00 | -7.90E+00 |
| | Sd. | 7.4E-03 | 3.3E+00 | **3.1E+00** | 3.74E+00 | 5.23E-01 | **1.90E+00** | 8.89E-04 | 9.93E-01 | 2.87E+00 |
| $F_{22}$ | Best | -1.0E+01 | **-1.0E+01** | -1.0E+01 | -1.04E+01 | -2.09E+00 | -1.04E+01 | -1.03E+00 | -4.83E+00 | **-1.04E+01** |
| | Avg. | -7.5E+00 | **-9.1E+00** | -1.0E+01 | -8.87E+00 | 3.63E-01 | -8.99E+00 | 3.00E+00 | -3.76+00 | **-1.04E+01** |
| | Sd. | 2.7E+00 | **2.8E+00** | 7.2E-15 | 3.14E+00 | 7.00E-01 | 2.35E+00 | 3.00E-01 | 1.22E+00 | **2.40E-05** |
| $F_{23}$ | Best | -1.1E+01 | -1.1E+01 | -1.1E+01 | -1.05E+01 | -2.06E+00 | -1.05E+00 | **-3.27E+00** | -4.81E+00 | **-1.05E+01** |
| | Avg. | -1.0E+01 | -9.4E+00 | -1.0E+01 | -7.9E+00 | 4.22E-01 | -9.40E+00 | **-1.04E+01** | -3.95E+00 | **-1.05E+01** |
| | Sd. | 7.8E-01 | 2.8E+00 | 1.6E+00 | 3.69E+00 | 9.31E-01 | 2.28E+00 | **1.05E+01** | 9.55E-01 | **4.76E-06** |

**Table 8:** Comparing average and standard deviation of proposed algorithms for unimodal functions with OBSCA, SCA, OBPSO, SSO.

| Function | Value Heads | **Proposed Algorithms** | | OBSCA | SCA | OBPSO | SSO |
|---|---|---|---|---|---|---|---|
| | | MGPS | MPSOGSA | | | | |
| $F_1$ | Best | **4.12E-21** | 3.10E-13 | **1.75E-75** | 6.89E-01 | 1.85E-07 | 1.65E-01 |
| | Avg. | **9.38E-19** | 1.92E-09 | **1.82E-74** | 5.43E+00 | 2.86E-06 | 1.90E-01 |
| | Sd. | **1.57E-18** | 3.67E-09 | **2.90E-11** | 1.54E+01 | 1.23E-05 | 8.47E-17 |
| $F_2$ | Best | **1.40E-20** | 2.11E-14 | **3.84E-45** | 1.25E-02 | 6.27E-03 | 1.00E+00 |
| | Avg. | **1.40E-20** | 1.09E-10 | **1.09E-42** | 2.37E-02 | 5.69E-02 | 2.05E+00 |
| | Sd. | **2.21E-20** | 4.48E-10 | **2.90E-11** | 4.35E-02 | 6.95E-02 | 9.03E-16 |
| $F_3$ | Best | **1.17E-06** | 2.5E-02 | 2.00E+00 | 3.65E+03 | 5.45E+00 | 1.12E+02 |
| | Avg. | **5.57E-07** | 9.98E-01 | 2.05E+01 | 1.02E+04 | 6.36E+01 | 1.14E+02 |
| | Sd. | **5.52E-06** | 7.1E+01 | 2.64E+00 | 6.38E+02 | 6.76E+01 | 2.89E-14 |
| $F_4$ | Best | **1.17E-06** | 8.30E-05 | **4.5E-34** | 2.65E+00 | 1.89E+00 | 1.75E+00 |
| | Avg. | **5.52E-06** | 3.19E-04 | **3.22E-32** | 3.63E+01 | 2.18E+00 | 2.09E+00 |
| | Sd. | **1.88E-06** | 2.74E-04 | **1.19E-01** | 1.38E+01 | 1.03E+00 | 9.03E-16 |
| $F_5$ | Best | **4.19E+00** | **2.24E+01** | 1.87E+00 | 5.54E+02 | 3.56E+01 | 6.65E+00 |
| | Avg. | **2.45E+01** | **2.51E+01** | 2.82E+01 | 6.30E+04 | 5.17E+01 | 4.54E+01 |
| | Sd. | **9.47E+00** | **6.65E-01** | 1.80E-01 | 1.98E+05 | 3.04E+01 | 1.45E-14 |
| $F_6$ | Best | 4.79E-06 | **1.75E-05** | 3.75E+00 | 2.78E+02 | **1.05E-08** | 1.45E-01 |
| | Avg. | 2.21E-02 | **1.26E-02** | 4.70E+00 | 6.30E+04 | **1.16E-06** | 1.91E-01 |
| | Sd. | 2.39E-02 | **5.59E-02** | 3.32E-01 | 1.49E+01 | **3.92E-06** | 8.47E-17 |
| $F_7$ | Best | 8.51E-04 | **1.13E-04** | **1.25E-05** | 1.08E-01 | 2.35E-03 | 2.78E-01 |
| | Avg. | 1.73E-02 | **1.6E-02** | **2.13E-04** | 1.35E-01 | 2.92E-02 | 3.82E-01 |
| | Sd. | 1.01E-02 | **1.0E-02** | **2.87E-03** | 1.60E-01 | 1.60E-02 | 2.26E-16 |

**Table 9:** Comparing average and standard deviation of proposed algorithms for multimodal functions with OBSCA, SCA, OBPSO, SSO.

| Function | Value Heads | **Proposed Algorithms** | | OBSCA | SCA | OBPSO | SSO |
|---|---|---|---|---|---|---|---|
| | | MGPS | MPSOGSA | | | | |
| $F_8$ | Best | **-9.11E+03** | -7.86E+03 | -6.85E+03 | -7.00E+03 | -7.24E+03 | **-9.89E+03** |
| | Avg. | **-7.46E+03** | -7.31E+03 | -3.53E+03 | -3.70E+03 | -6.06E+03 | **-8.90E+03** |
| | Sd. | **6.21E+02** | 4.76E+02 | 2.74E+02 | 3.03E+02 | 1.00E+03 | **5.55E-12** |
| $F_9$ | Best | 1.14E-13 | **1.29E-11** | **0.00E+00** | 3.45E+01 | 2.45E+00 | 6.75E+01 |
| | Avg. | 3.47E+00 | **2.89E-07** | **0.00E+00** | 4.90E+01 | 4.58E+01 | 7.53E+01 |
| | Sd. | 2.84E+00 | **1.06E-06** | **2.08E-09** | 4.06E+01 | 1.35E+01 | 1.45E-14 |
| $F_{10}$ | Best | **2.86E-11** | 5.62E-07 | **3.84E-13** | 4.56E+00 | 1.15E+00 | 3.65E-01 |
| | Avg. | **1.42E-08** | 5.83E-05 | **8.88E-16** | 1.66E+01 | 1.52E+00 | 4.85E-01 |
| | Sd. | **3.59E-08** | 5.28E-05 | **2.07E+00** | 7.12E+00 | 8.02E-01 | 3.95E-16 |
| $F_{11}$ | Best | **0.0E+00** | **6.79E-14** | 2.25E-02 | 7.75E-02 | 3.26E-03 | 1.04E-02 |
| | Avg. | **8.49E-05** | **1.09E-11** | 1.00E-01 | 8.88E-01 | 2.69E-02 | 1.04E-02 |
| | Sd. | **2.96E-04** | **1.75E-11** | 7.00E-02 | 3.13E-01 | 3.84E-02 | 0.00E+00 |
| $F_{12}$ | Best | **1.00E-03** | **2.66E-08** | 4.73E-01 | 1.45E+04 | 1.54E-02 | 2.56E+00 |
| | Avg. | **1.9E-03** | **1.2E-02** | 5.72E-01 | 2.89E+04 | 1.56E-01 | 3.27E+00 |
| | Sd. | **5.32E-04** | **2.4E-03** | 1.80E-01 | 1.07E+05 | 2.85E-01 | 4.52E-16 |
| $F_{13}$ | Best | **2.56E-02** | 5.93E-01 | 1.75E+00 | 7.8E+03 | 3.45E-02 | **1.14E+01** |
| | Avg. | **4.33E-01** | 9.73E-01 | 2.41E+00 | 6.75E+04 | 7.83E-02 | **1.14E-01** |
| | Sd. | **3.1E-01** | 1.88E-01 | 1.69E-01 | 1.98E+05 | 2.00E-01 | **0.00E+00** |

**Table 10:** Comparing average and standard deviation of proposed algorithms for multimodal functions with a fixed dimension, with OBSCA, SCA, OBPSO, SSO.

| Function | Value Heads | **Proposed Algorithms** | | OBSCA | SCA | OBPSO | SSO |
|---|---|---|---|---|---|---|---|
| | | MGPS | MPSOGSA | | | | |
| $F_{14}$ | Best | **1.0E+00** | **1.0E+00** | 1.37E+00 | 2.09E+00 | 1.07E+00 | 1.45E+00 |
| | Avg. | **1.0E+00** | **1.51E+00** | 2.64E+00 | 2.18E+00 | 3.40E+00 | 2.98E+00 |
| | Sd. | **2.84E-11** | **6.81E-01** | 3.11E+00 | 2.49E+00 | 2.74E+00 | 4.52E-16 |
| $F_{15}$ | Best | 3.53E-04 | **3.23E-04** | 4.56E-04 | 2.89E-01 | 2.34E-04 | **3.45E-04** |
| | Avg. | 9.05E-04 | **3.59E-04** | 6.58E-04 | 1.08E+00 | 1.88E-03 | **7.45E-04** |
| | Sd. | 4.43E-14 | **4.07E-05** | 2.83E-04 | 3.78E-04 | 5.04E-03 | **3.31E-19** |
| $F_{16}$ | Best | **-1.03E+00** | **-1.03E+00** | -1.03E+00 | -1.03E+00 | -1.03E+00 | -1.03E+00 |
| | Avg. | **-1.03E+00** | **-1.03E+00** | -1.05E+00 | -1.01E+00 | -1.02E+00 | -1.04E+00 |
| | Sd. | **0.00E+00** | **6.83E-17** | 8.51E-06 | 4.46E-05 | 6.45E-16 | 9.06E-16 |
| $F_{17}$ | Best | **3.98E-01** | **3.98E-01** | 3.98E-01 | 3.99E-01 | **3.98E-01** | **3.98E-01** |
| | Avg. | **3.98E-01** | **3.98E-01** | 3.99E-01 | 4.00E-01 | **3.98E-01** | **3.98E-01** |
| | Sd. | **1.49E-05** | **0.00E+00** | 6.55E-04 | 1.43E-03 | **0.00E+00** | **0.00E+00** |
| $F_{18}$ | Best | **3.00E+00** | 3.00E+00 | 3.00E+00 | 3.00E+00 | 3.00E+00 | **3.00E+00** |
| | Avg. | **3.00E+00** | 3.01E+00 | 3.10E+00 | 3.13E+00 | 3.24E+00 | **3.00E+00** |
| | Sd. | **6.09E-16** | 9.38E-15 | 6.54E-05 | 1.56E-04 | 1.22E-15 | **0.00E+00** |
| $F_{19}$ | Best | **-3.85E+00** | **-3.88E+00** | -3.81E+01 | -3.56E-01 | -3.45E-01 | -2.86E-01 |
| | Avg. | **-3.87E+00** | **-3.88E+00** | -3.00E-01 | -3.00E-01 | -3.00E-01 | -2.86E-01 |
| | Sd. | **2.70E-02** | **1.63E-16** | 2.26E-16 | 2.26E-16 | 2.26E-16 | 0.00E+00 |
| $F_{20}$ | Best | -3.32E+00 | **-3.32E+00** | -3.25E+00 | -3.20E+00 | -3.29E+00 | **-3.31E+00** |
| | Avg. | -3.26E+00 | **-3.32E+00** | -3.10E+00 | -3.04E+00 | -3.29E+00 | **-3.31E+00** |
| | Sd. | 4.75E-02 | **0.00E+00** | 3.94E-02 | 1.16E-01 | 5.35E-02 | **0.00E+00** |
| $F_{21}$ | Best | -4.99E+00 | -1.02E+01 | **-1.04E+01** | -6.57E+01 | -7.65E+01 | **-1.05E+01** |
| | Avg. | -3.61+00 | -7.90E+00 | **-9.06E+00** | -2.20E+00 | -6.24E+00 | **-9.49E+00** |
| | Sd. | 9.93E-01 | 2.87E+00 | **1.76E+00** | 1.71E+00 | 3.74E+00 | **3.61E-15** |
| $F_{22}$ | Best | -4.83E+00 | **-1.04E+01** | -9.95E+00 | -6.78E+00 | -9.95E+00 | **-1.04E+01** |
| | Avg. | -3.76+00 | **-1.04E+01** | -9.93E+00 | -4.27E+00 | -8.33E+00 | **-1.04E+01** |
| | Sd. | 1.22E+00 | **2.40E-05** | 2.73E-01 | 1.43E+00 | 3.26E+00 | **5.42E-15** |
| $F_{23}$ | Best | -4.81E+00 | **-1.05E+01** | -1.02E+01 | -6.67E+00 | -9.87E+00 | **-1.05E+01** |
| | Avg. | -3.95E+00 | **-1.05E+01** | -1.01E+01 | -3.34E+00 | -8.21E+00 | **-1.05E+01** |
| | Sd. | 9.55E-01 | **4.76E-06** | 2.56E-01 | 1.78E+00 | 3.41E+00 | **0.00E+00** |

## 4.2 Graphical depiction of the Results

We use graphs to show the convergence of our algorithms for some of the benchmark functions considered here. These graphs are plotted using the values for the best-obtained fitness value in the population for each iteration of the algorithm. For all the plots *gbest* refers to the global best fitness value for all members in the population. For $F_{15}$, 0.00030 is added to the fitness result before plotting and for $F_{19}$, 3.8774 is added. We have shown one graph from each of the three categories of functions.

Figure 3 (A, C and E; B, D and F) shows the variation of the *gbest* value with the iterations for MGPS and MPSOGSA respectively. *gbest* denotes the fitness of the best particle of the population. The graphs show the convergence of the population. The spikes are due to mutation which causes variations in the global fitness to prevent the method from getting stuck in local optima. Some functions have wide fluctuation in performance like $F_9$ and $F_{19}$. It should be noted that in Tables 8-10 the performance of the proposed algorithm for $F_{19}$ is the best and for $F_9$ is second best. Similarly, for Tables 6-8 the performance for $F_{19}$ is second best for the proposed algorithms. So, it can be concluded that the fluctuation does not hamper performance but is rather helpful in avoidance of local minima and premature convergence.

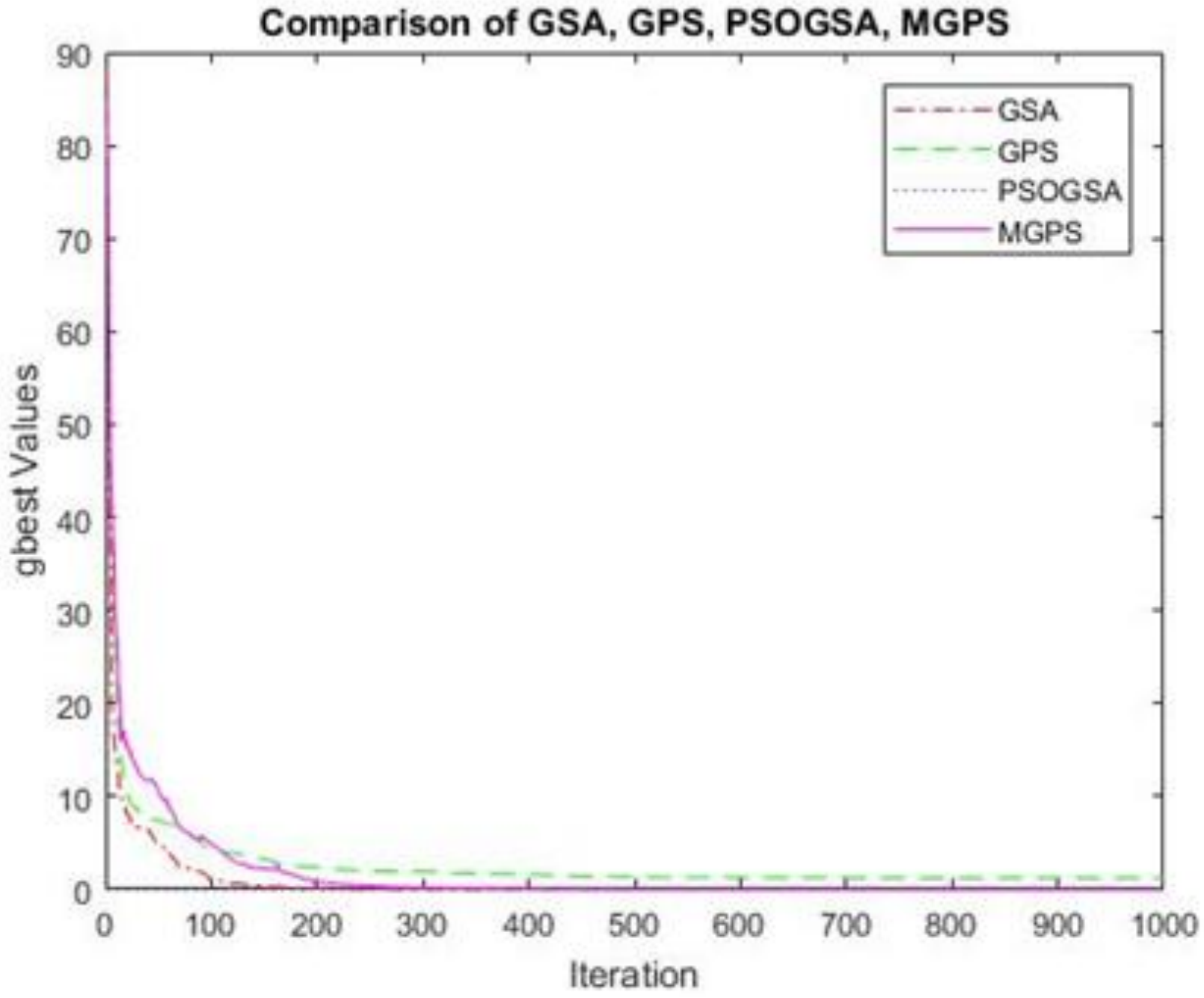

**A:** Graph comparing $gbest$ of function $F_4$ at each iteration using GSA, GPS, PSOGSA and MGPS algorithms

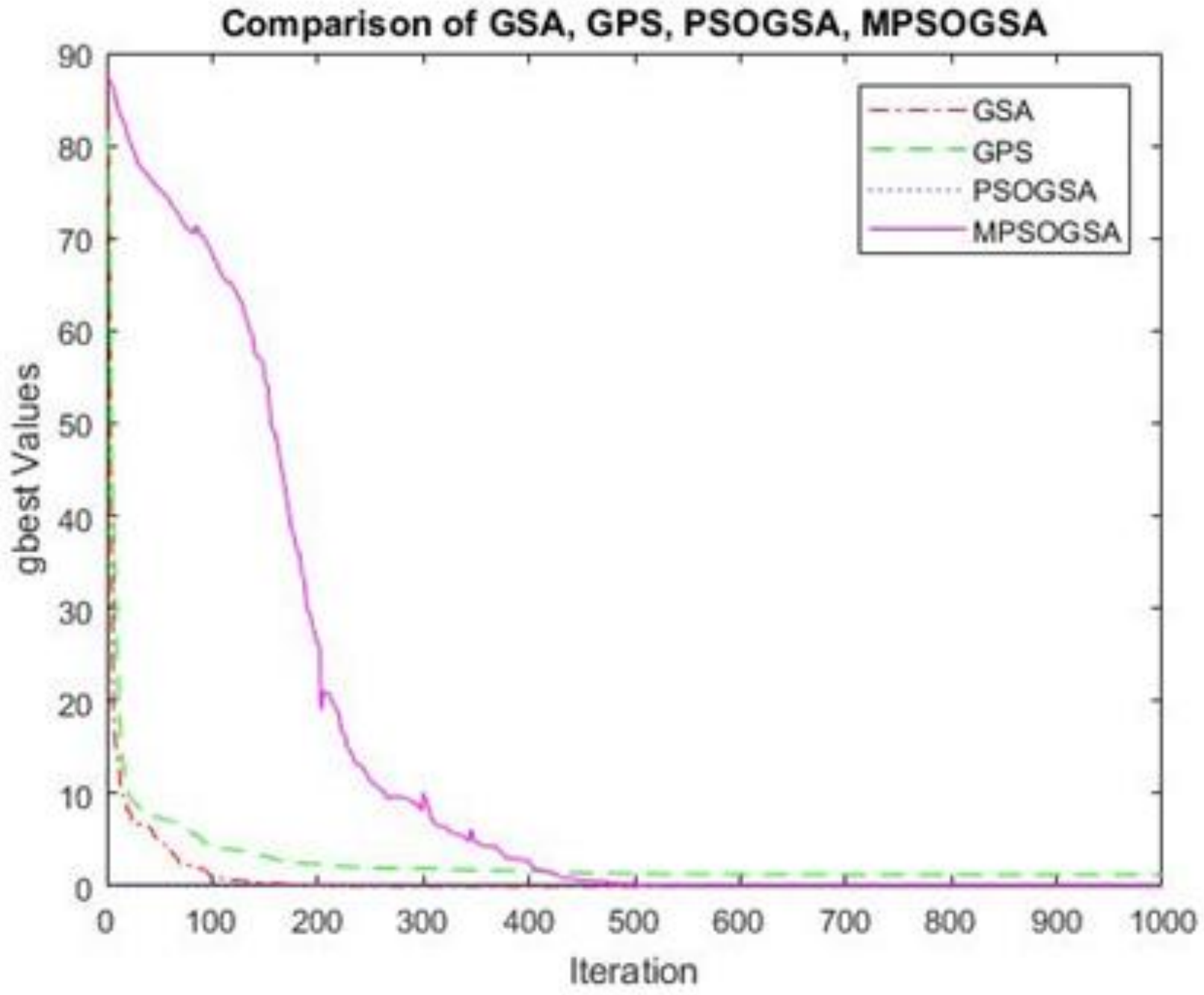

**B:** Graph comparing $gbest$ of function $F_4$ at each iteration using GSA, GPS, PSOGSA and MPSOGSA algorithms

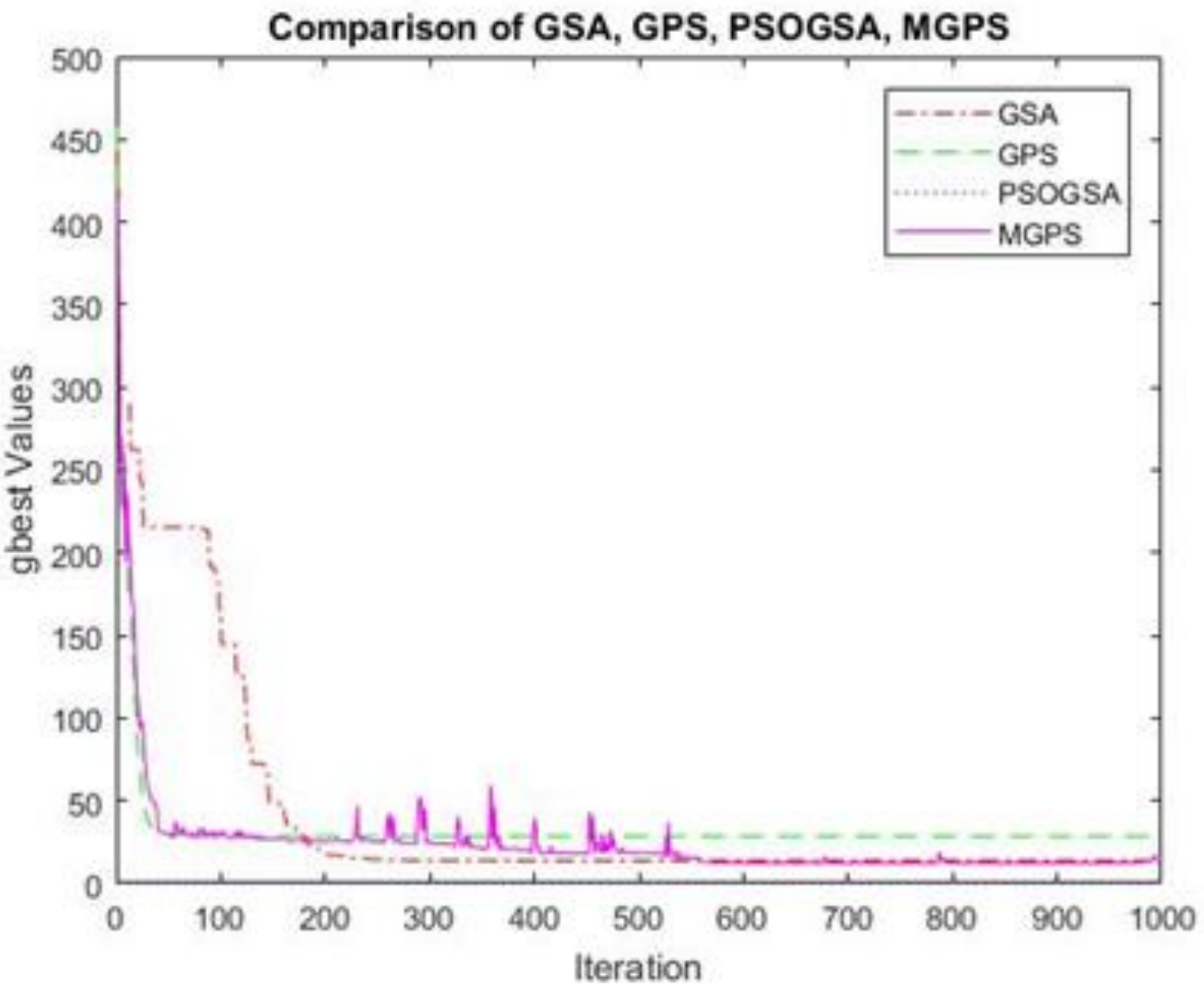

**C:** Graph comparing $gbest$ of function $F_9$ at each iteration using GSA, GPS, PSOGSA and MGPS algorithms

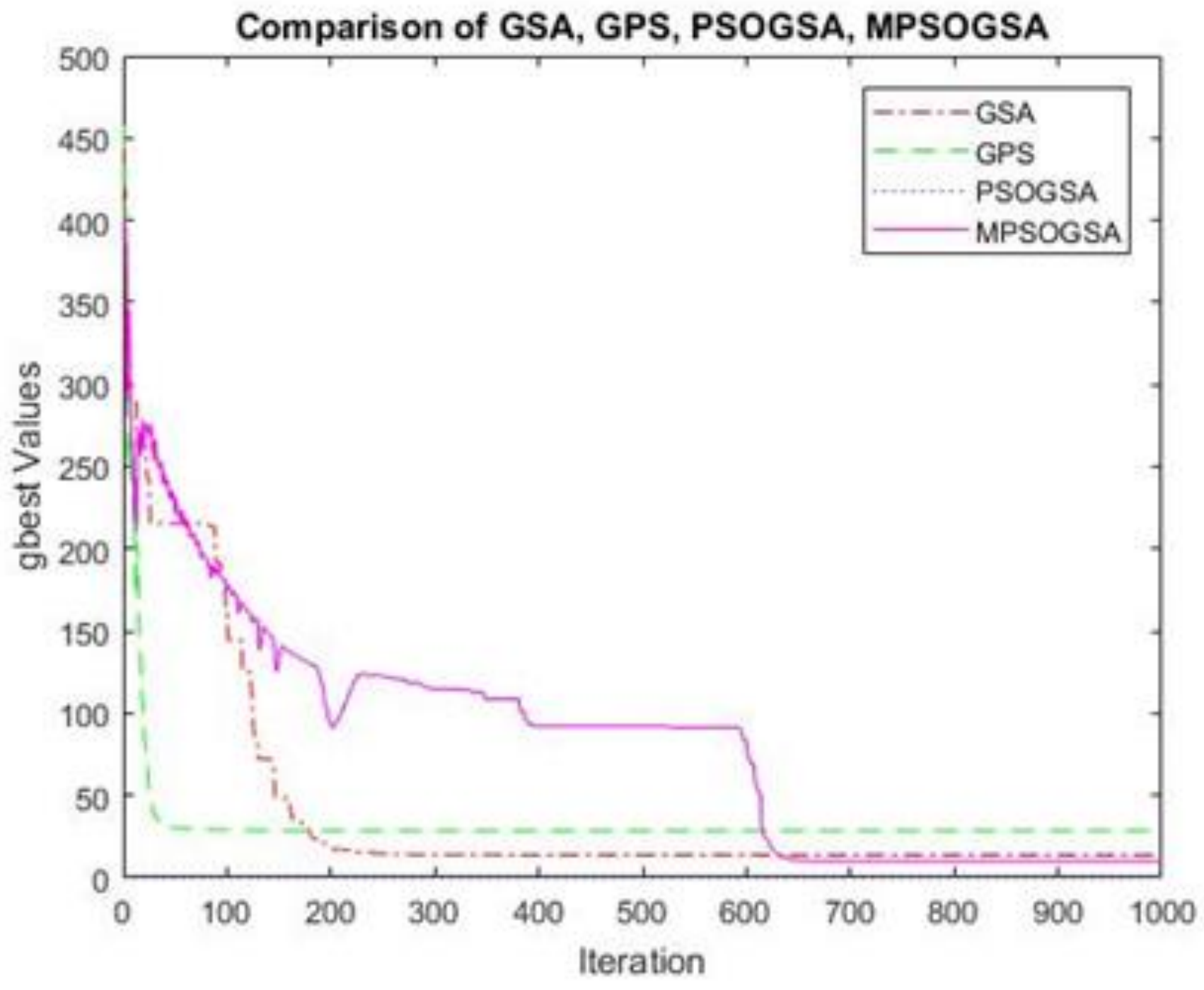

**D:** Graph comparing $gbest$ function $F_9$ at each iteration using GSA, GPS, PSOGSA and MPSOGSA algorithms

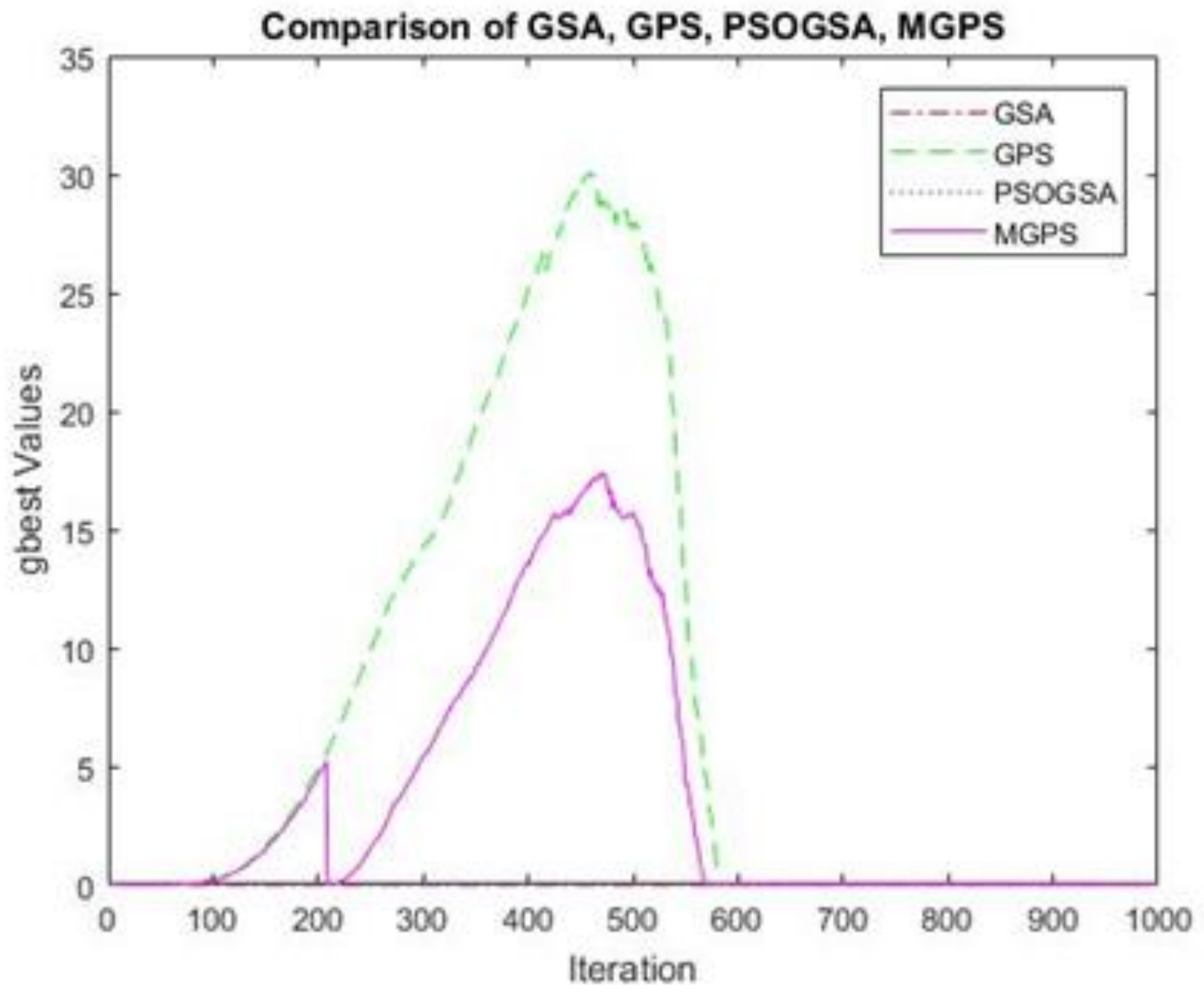

**E:** Graph comparing $gbest$ function $F_{15}$ at each iteration using GSA, GPS, PSOGSA and MGPS algorithms

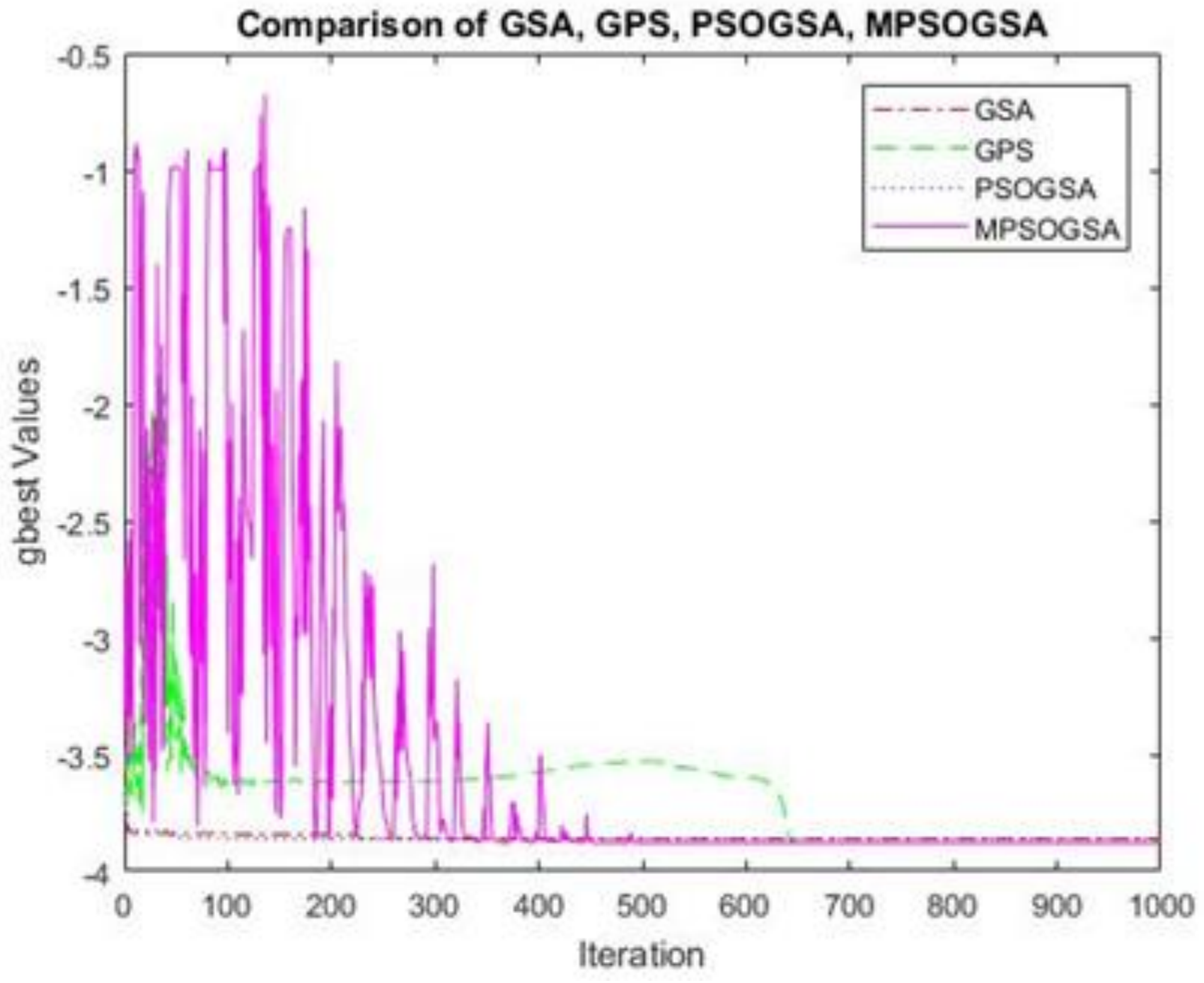

**F:** Graph comparing $gbest$ of function $F_{19}$ at each iteration using GSA, GPS, PSOGSA and MPSOGSA algorithms

**Figure 3 (A-F)**: Variations of the value of $gbest$ of the particles in the population versus iterations for MGPS/ MPSOGSA, along with their ancestors (GPS and PSOGSA), PSO and GSA. A, C, E- contain the plots of the MGPS algorithm and B, D, F contain the plots of MPSOGSA algorithm.

### 4.3 Time requirement analysis

From the results and corresponding discussion provided in sections 4.1 and 4.2, it is clear that the addition of mutation does help PSOGSA and GPS avoid premature convergence, thereby helping them to search for a global optimum solution. But, how computationally expensive mutation operation is? To find that out, we have provided the time requirements of both the algorithms before and after adding mutation operation over the 23 benchmark functions. The graphical representations of the time requirements are shown in Figure 4 (A, B).

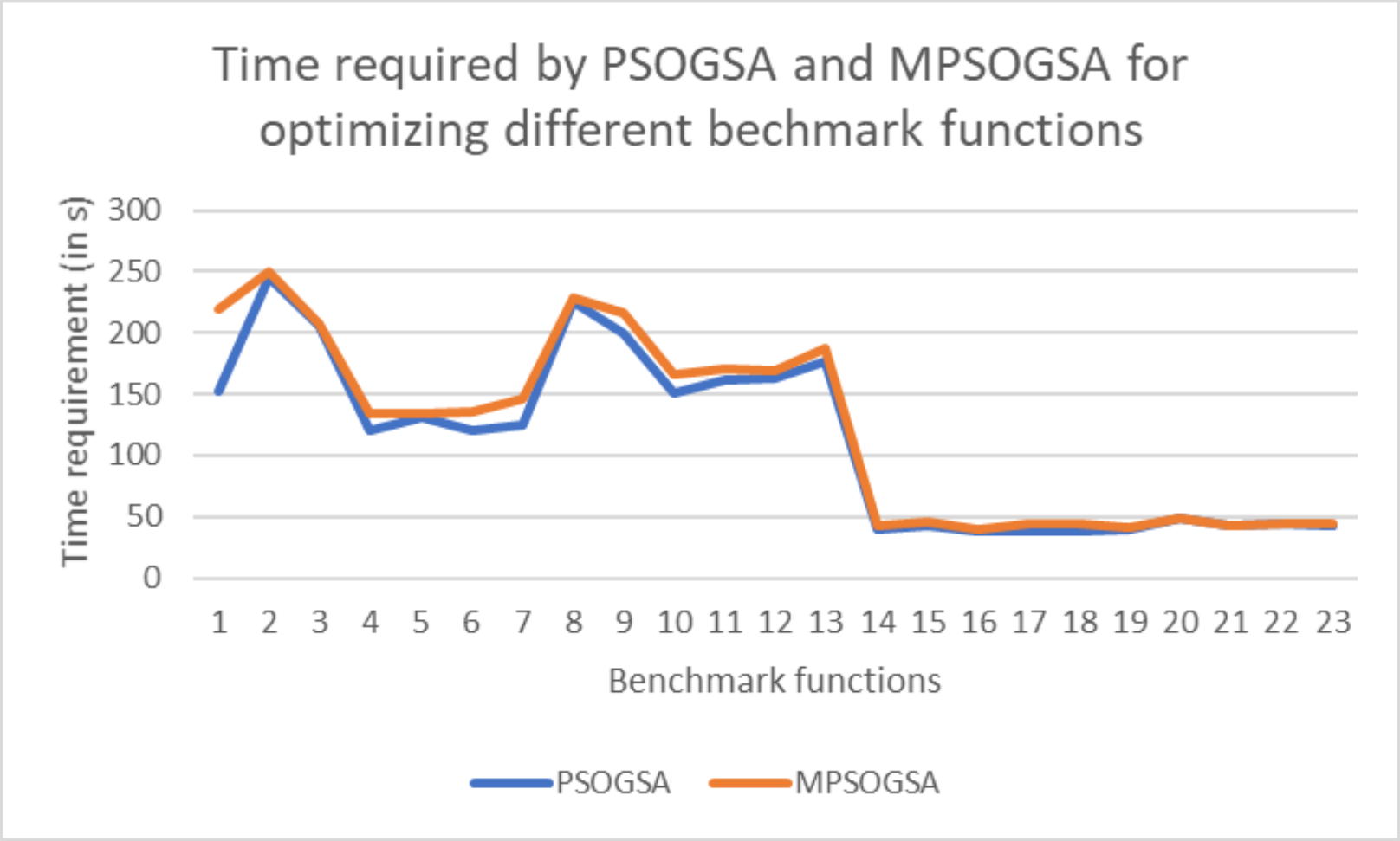

**A:** Graph representing time requirement against 23 benchmark functions for PSOGSA and MPSOGSA.

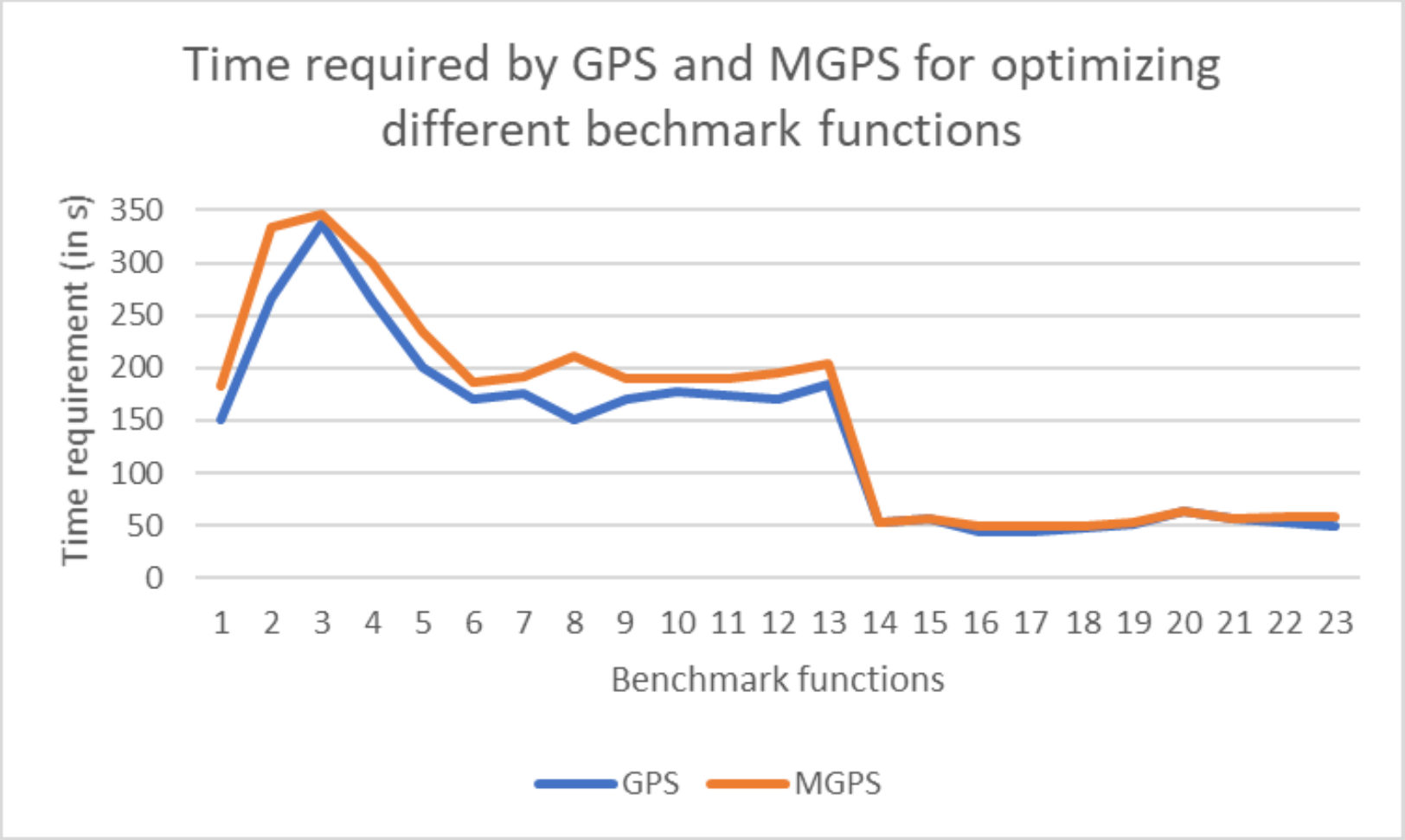

**B:** Graph representing time requirement against 23 benchmark functions for GPS and MGPS.

**Figure 4 (A-B):** The time required by PSOGSA and GPS before and after addition of mutation operation for optimizing benchmark functions.

From the Fig. 4, it is clear that the mutation operation requires very small amount of time to guide PSOGSA and GPS to a better solution. For function 13 onwards, the time required by MPSOGSA or MGPS almost coincides with that of PSOGSA and GPS respectively. For functions 1-12, there is a small increase in the time requirement but considering the improvement in the results, this amount of time increase is quite insignificant.

## 5. Applications of the proposed algorithm to solve Engineering Design Problems

Engineering design involves building and designing of products and/or processes. It is a decision-making process requiring complex objective function optimization. Meta-heuristic methods (Simulated Annealing or Tabu search [50]) serve as a better approach than traditional optimization methods like random walk, exhaustive search or steepest descent method. Meta-heuristics converge to an optimal solution and can handle non-convex and non-differentiable functions. Engineering design problems have a large number of variables, whereas their influence on the objective function can be very complicated. Therefore, in this paper, five classical engineering design problems *viz*. spring, gear train, welded beam, pressure vessel and closed coil helical spring design have been considered which are discussed in sections 5.1, 5.2, 5.3, 5.4 and 5.5 respectively. These problems contain various local optima, whereas only global optimum is required. Hence, there is a need for effective and efficient optimization methods for them. In this section, various experiments on these benchmark problems are reported to verify the performance of the proposed algorithms. All the experiments are performed over 30 independent runs for 1000 iterations. The constraints for the problems can be found in [36]. The results for each of the functions have been shown in Table 11 and they have been compared with the hybrid algorithms, GPS and PSOGSA. The respective variable values for each of the functions are given in Table 12. From the comparison of the results obtained over benchmark functions presented in Tables 5-10, we can see that VPL and LCA are two algorithms which have performed quite well in the scenario. That is why we have also applied them over engineering design problems and compared their results with MBPSOGSA and MGPS in Table 11. Following are specific engineering problems which can be solved by using the proposed algorithms efficiently.

### 5.1. Tension/Compression Spring Design Problem

The weight of the spring is based on three decision variables, *namely*, the wire diameter (d), mean coil diameter (D) and the number of active coils (N). The weight is minimized subjected to three inequality constraints and the objective function in Eq. 24. A population particle is given as, $\vec{x} = [x_1 x_2 x_3] = [dDN]$.

$$EF_1 = (x_3 + 2)x_2 x_1^2 \qquad (24)$$

### 5.2. Gear Train Design Problem

Here the cost of gear ratio, of the gear train, is minimized. The problem has no equality or inequality constraint except a boundary constraint. It consists of four decision variables $n_A(x_1)$ , $n_B(x_2)$, $n_D(x_3)$ , $n_F(x_4)$ using which the gear ratio can be formulated as $n_B n_D / n_F n_A$. The objective function to be minimized is,

$$EF_2(x) = \left((1/6.931) - (x_3 x_2 / x_1 x_4)\right)^2$$

Subject to, $12 \le x_i \le 60$ (25)

### 5.3. Welded Beam Design Problem

This is a minimization problem having four variables namely weld thickness $(h)$, length of the bar attached to the weld $(l)$, bar's height $(t)$, and bar's thickness $(b)$. The constraints for this problem include bending stress $(\theta)$, bean deflection $(\delta)$, shear stress $(\tau)$, buckling load $(P_c)$ and other constraints. The population point is taken as $\vec{x} = [x_1 x_2 x_3 x_4] = [hltb]$. The objective function is,

$$EF_3(x) = 1.1047x_1^2x_2 + 0.04811x_3x_4(14.0 + x_2)$$
(26)

**5.4. Pressure Design Vessel Problem**

Pressure vessel design problem involves minimization of the welding, manufacturing and material cost of the pressure vessel. There are four decision variables involved in this problem which are the thickness of shell ($T_S$), the thickness of head ($T_h$) which are discrete decision variables, inner radius ($R$) and length of the cylindrical section of the vessel ($L$) which are continuous decision variables. The population point is taken as $\vec{x} = [x_1 x_2 x_3 x_4] = [T_S T_h R L]$. The objective function is,

$$EF_4(\vec{x}) = 0.6224x_1x_3x_4 + 1.7781x_2x_3^2 + 3.1661x_1^2x_4 + 19.84x_1^2x_3 \quad (27)$$

**5.5. Closed Coil Helical Spring Design Problem**

The volume of the closed coil helical spring is minimized. Helical spring is made up of closed coil wire having the shape of a helix and is intended for the tensile and compressive load. The population point is given as $\vec{x} = [x_1 x_2 x_3] = [dDN_c]$. There are chiefly two decision variables to consider namely coil diameter(D) and wire diameter(d). The number of coils ($N_c$) can be fixed beforehand. The volume of the helical spring (U) is given as the minimization function,

$$EF_5 = \frac{\pi^2}{4}(N_c + 2)Dd^2 \quad (28)$$

**Table 11**: Function values for the design problems - spring, gear train, welded beam, pressure vessel and closed coil helical spring design problem.

| Function | Value Heads | GPS | PSOGSA | VPL | LCA | MGPS | MPSOGSA |
|---|---|---|---|---|---|---|---|
| $EF_1$ | Best<br>Avg.<br>Sd. | 6.9E-03<br>1.07E-01<br>2.5E-03 | 2.5E-03<br>2.5E-03<br>8.64E-10 | 1.24E-3<br>2.37E-2<br>1.8E-03 | 1.26E-3<br>2.21E-2<br>3.88E-3 | **2.5E-03**<br>**2.5E-03**<br>**8.90E-19** | **2.5E-03**<br>**2.5E-03**<br>**8.90E-19** |
| $EF_2$ | Best<br>Avg.<br>Sd. | 2.88E-07<br>2.00E-03<br>4.0E-03 | 3.6E-03<br>2.00E-03<br>4.93E-09 | 2.8E−12<br>2.5E-09<br>3.9E-06 | 2.5E-11<br>3.8E-08<br>1.1E-09 | **5.01E-14**<br>**8.34E-04**<br>**1.80E-03** | **0.00E+00**<br>**0.00E+00**<br>**0.00E+00** |
| $EF_3$ | Best<br>Avg.<br>Sd. | **0.00E+00**<br>**0.00E+00**<br>**0.00E+00** | **0.00E+00**<br>**0.00E+00**<br>**0.00E+00** | 2.26E+0<br>3.21E+0<br>4.7E-16 | 1.72E+0<br>1.72E+0<br>7.1E-15 | **0.00E+00**<br>**0.00E+00**<br>**0.00E+00** | **0.00E+00**<br>**0.00E+00**<br>**0.00E+00** |
| $EF_4$ | Best<br>Avg.<br>Sd. | 0.00E+00<br>8.88E+03<br>4.48E+04 | 0.00E+00<br>3.33E+02<br>4.79E+02 | 6.04E+3<br>6.87E+3<br>1.32E+1 | 6.06E+3<br>6.07E+3<br>11.4E+0 | **0.00E+00**<br>**0.00E+00**<br>**0.00E+00** | **0.00E+00**<br>**0.00E+00**<br>**0.00E+00** |
| $EF_5$ | Best<br>Avg.<br>Sd. | **13.75E+00**<br>**13.75E+00**<br>**0.00E+00** | **13.75E+00**<br>**13.75E+00**<br>**0.00E+00** | 40.1E+0<br>41.9E+0<br>2.3E+00 | 42.8E+0<br>43.7E+0<br>1.7E+00 | **13.75E+00**<br>**13.75E+00**<br>**0.00E+00** | **13.75E+00**<br>**13.75E+00**<br>**0.00E+00** |

**Table 12:** Value of the parameters after optimization.

| Function | Algorithm | $x_1$ | $x_2$ | $x_3$ | $x_4$ | $f(x)$ |
|---|---|---|---|---|---|---|
| $EF_1$ | MGPS | 0.05 | 0.25 | 2.00 | N/A | 2.5E-03 |
| | MPSOGSA | 0.05 | 0.25 | 2.00 | N/A | 2.5E-03 |
| $EF_2$ | MGPS | 60 | 12 | 43.2837 | 60 | 0 |
| | MPSOGSA | 29.2062 | 12 | 12.0749 | 34.3865 | 0 |
| $EF_3$ | MGPS | 0.1 | 0.1 | 0.1 | 0.1 | 0 |
| | MPSOGSA | 0.1 | 0.1 | 0.1 | 0.1 | 0 |
| $EF_4$ | MGPS | 0 | 0 | 82.7991 | 10.6423 | 0 |
| | MPSOGSA | 0 | 0 | 79.0777 | 10 | 0 |
| $EF_5$ | MGPS | 0.508 | 1.27 | 15 | N/A | 13.74738 |
| | MPSOGSA | 0.508 | 1.27 | 15 | N/A | 13.74738 |

**Note:** In case of $EF_4$ in 5.4 since the lower boundary for both $x_1$ and $x_2$ is 0 the algorithm converges towards 0.

## 6. Conclusion and Future Work

The hybrids of GSA and PSO – GPS and PSOGSA are found to be efficient in single-objective optimization but suffer from premature convergence. This problem is addressed in the present work by the use of mutation and hence better optimizations results are obtained. We have proposed a model of fuzzy mutation based on the distances between the points from the centroid and the population history, which outperforms their ancestors GPS and PSOGSA. The evaluation of the models on benchmark functions provides impressive results. To show the practical application of our proposed algorithms, the same has been evaluated on five classic engineering design problems. The results are quite promising and our algorithms outperform their ancestors in most cases or are shoulder to shoulder. This model of mutation is not algorithm-specific and can be applied to any algorithm which suffers from premature convergence like Whale optimization or Harmony search algorithm. Future scope of this work might involve the use of a local and a global change counters to perform mutation.

## Funding Information

Acknowledgment: This work has not received any funds.

Authors have no competing interests.

**Compliance with Ethical Standards:**

Conflict of Interest: Authors have no Conflict of Interest.

Ethical approval: This article does not contain any studies with human participants or animals performed by any of the authors.